%% file: main.tex
\documentclass{jingdong}
\input{preamble}

\title{GUI-JEV: Grounded Action Selection for Fast GUI Automation}

\author[1,2,3,*]{Shuai Dong}
\author[3,4,*]{Ziyue Wang}
\author[2,3,*]{Ji Zhao}
\author[3,*]{Yongfu Zhu}
\author[3,5,*]{Wenzheng Chang}
\author[3,6,*]{Caoyuan Ma}
\author[7,*]{Haiyuan Wan}
\author[3,8]{Taiqiang Wu}
\author[3]{Xuehui Wang}
\author[1,3,5]{Kanghui Tian}
\author[9]{Xiangnan Wu}
\author[3]{Congcong Wang}
\author[3,10]{Kaiwen Tuo}
\author[3,11]{Hongrui Wu}
\author[12]{Siyuan Wang}
\author[3,\dagger]{Wenqi Shao}
\author[1,2,\dagger]{Zhongyu Wei}
\author[3,\ddagger]{Jiaqi Wang}

\affiliation[1]{Fudan University}
\affiliation[2]{Shanghai Innovation Institute}
\affiliation[3]{JD.com}
\affiliation[4]{Peking University}
\affiliation[5]{Shanghai Artificial Intelligence Laboratory}
\affiliation[6]{The University of Tokyo}
\affiliation[7]{Tsinghua University}
\affiliation[8]{The University of Hong Kong}
\affiliation[9]{Institute of Automation, Chinese Academy of Sciences}
\affiliation[10]{The Hong Kong University of Science and Technology}
\affiliation[11]{Stanford University}
\affiliation[12]{The Chinese University of Hong Kong}

\contribution[*]{Equal contribution}
\contribution[\ddagger]{Project leader}
\contribution[\dagger]{Corresponding authors.}

\resources{%
  {\fontsize{12}{14}\selectfont\sffamily\bfseries
  \mbox{\resourcecodeicon\hspace{0.4em}\href{https://github.com/XD111ds/GUI-JEV-REPO}{Code}}%
  \hspace{2em}%
  \mbox{\resourcemodelicon\hspace{0.4em}\href{https://huggingface.co/shuai22/gui-jev}{Model}}%
  \hspace{2em}%
  \mbox{\resourcedataicon\hspace{0.4em}\href{https://github.com/XD111ds/GUI-JEV-REPO}{Data}}}
}

\hypersetup{
    pdftitle={GUI-JEV: Grounded Action Selection for Fast GUI Automation},
    pdfauthor={Shuai Dong, Ziyue Wang, Ji Zhao, Yongfu Zhu, Wenzheng Chang, Caoyuan Ma, Haiyuan Wan, Taiqiang Wu, Xuehui Wang, Kanghui Tian, Xiangnan Wu, Congcong Wang, Kaiwen Tuo, Hongrui Wu, Siyuan Wang, Wenqi Shao, Zhongyu Wei, Jiaqi Wang},
    pdfsubject={Preprint}
}

\abstract{\input{sec/0_abstract}}

\begin{document}
\maketitle
\input{sec/teaser}
\input{sec/1_intro}
\input{fig/architecture_float}
\input{sec/2_related_work}
\FloatBarrier
\input{sec/3_methodology}
\FloatBarrier
\input{sec/data_flywheel}
\clearpage
\input{sec/benchmark}

\clearpage
\input{sec/4_experiments}

\suppressfloats[t]
\input{sec/5_results_and_discussions}
\FloatBarrier
\input{sec/6_conclusion}
\par
\noindent\textbf{LLM Usage Statement.}
We primarily used ChatGPT to polish the wording and improve its clarity and readability. Its use in the manuscript focused on language refinement and presentation. We also used it to help write code. The authors are responsible for reviewing the AI-assisted material and for the final claims, analyses, and conclusions.
\par

\FloatBarrier

\input{sec/references}
\clearpage
\input{sec/appendix}
\end{document}

%% file: preamble.tex
\usepackage{amsmath,amssymb}
\usepackage{array}
\usepackage{colortbl}
\usepackage[shortlabels,inline]{enumitem}
\usepackage{xspace}
\usepackage{wrapfig}
\usepackage{tabularx,longtable}
\definecolor{cvprblue}{rgb}{0.21,0.49,0.74}

\newcommand{\resourcecodeicon}{%
  \tikz[baseline=-0.12em,x=0.85em,y=0.85em,line width=0.6pt,line cap=round,line join=round]{%
    \draw (0.25,0.15) -- (0,0.4) -- (0.25,0.65);
    \draw (0.75,0.15) -- (1,0.4) -- (0.75,0.65);
    \draw (0.4,0) -- (0.6,0.8);
  }}
\newcommand{\resourcemodelicon}{%
  \tikz[baseline=-0.12em,x=0.85em,y=0.85em,line width=0.6pt,line join=round]{%
    \draw (0.45,0) -- (0.9,0.25) -- (0.9,0.75) -- (0.45,1) -- (0,0.75) -- (0,0.25) -- cycle;
    \draw (0,0.75) -- (0.45,0.5) -- (0.9,0.75);
    \draw (0.45,0.5) -- (0.45,0);
  }}
\newcommand{\resourcedataicon}{%
  \tikz[baseline=-0.12em,x=0.85em,y=0.85em,line width=0.6pt,line join=round]{%
    \draw (0.45,0.75) ellipse [x radius=0.45,y radius=0.15];
    \draw (0,0.75) -- (0,0.15) arc [start angle=180,end angle=360,x radius=0.45,y radius=0.15] -- (0.9,0.75);
    \draw (0,0.45) arc [start angle=180,end angle=360,x radius=0.45,y radius=0.15];
  }}

\renewcommand{\authorfont}{\fontsize{11}{13}\selectfont}

\definecolor{resultgreen}{RGB}{200,229,179}
\newcommand{\method}{GUI-JEV\xspace}

\crefname{figure}{Fig.}{Figs.}
\Crefname{figure}{Fig.}{Figs.}
\crefname{table}{Tab.}{Tabs.}
\Crefname{table}{Tab.}{Tabs.}

%% file: sec/0_abstract.tex
Everyday \emph{type-free} workflows, such as organizing files and adjusting settings, require interface understanding, progress tracking, and constraint-aware decisions. Strong GUI agents often rely on explicit reasoning for these decisions, adding decoding latency at each interaction. Yet type-free GUI execution uses fixed operation types and structured parameters.
GUI-JEV turns this structured interface into single-pass grounded control, predicting executable action blocks without generating explicit reasoning traces.
A shared visual pointer with within-patch offset refinement grounds targets directly in screenshots, while specialized heads predict operations and finite parameters.
A replay-verified data flywheel provides demonstrations validated by native task evaluation and exact replay on fresh desktops.
We further introduce OSWorld-JEV, a 129-task benchmark evaluating grounded action selection and constraint-aware execution across six categories of everyday \emph{type-free} workflows.
With a 35B-total, 3B-active backbone, GUI-JEV leads the evaluated models on ScreenSpot-v2 and ScreenSpot-Pro at 97.72 and 79.95.
On OSWorld-JEV, it reaches 48.84 at 1.65\,s per step, exceeding Qwen-CUA by 5.44 points with a $4.30\times$ relative speedup.
Composed with an on-demand writer, the same policy reaches 62.33 on OSWorld-Verified.

%% file: sec/teaser.tex
\begin{figure}[!t]
  \centering
  \includegraphics[width=\linewidth]{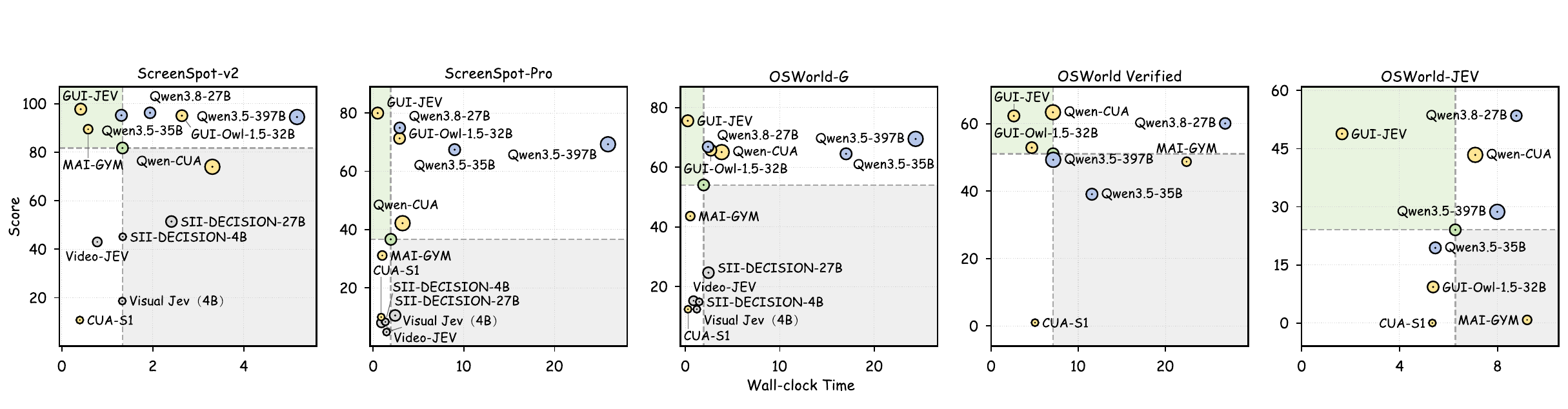}
  \caption{\textbf{Strong grounding and GUI control at low latency.}
  GUI-JEV combines leading ScreenSpot-v2 and ScreenSpot-Pro accuracy among evaluated models with the lowest reported timing on both interactive benchmarks.
  On OSWorld-JEV, it improves completion over Qwen-CUA by 5.44 percentage points at $4.30\times$ relative speed.
  Each panel plots benchmark score against time. Higher scores and lower times are better.
  On OSWorld-JEV, non-Jev baselines are allowed TYPE, while GUI-JEV runs type-free.}
  \label{fig:efficiency}
\end{figure}

%% file: sec/1_intro.tex
\section{Introduction}
\label{sec:introduction}

Everyday computer tasks such as organizing files, comparing products, adjusting settings, and processing business records often admit \emph{type-free} workflows that manipulate existing content without typing new text.
These workflows require agents to interpret visible content, maintain task constraints, and decide both \emph{what to do} and \emph{where to act}.
They therefore call for precise and efficient graphical user interface (GUI) control grounded in the visual state of the application.

Recent multimodal agents such as UI-TARS, GUI-Owl-1.5, and Qwen-CUA map instructions and screenshots to computer operations~\cite{qin2025uitars,xu2026guiowl,lu2026qwencua}.
These agents often use explicit reasoning to interpret interface states, track task progress, and select actions under task constraints. However, generating lengthy reasoning traces before each interaction adds substantial decoding latency. Despite this decision complexity, type-free GUI execution follows a structured interface with a fixed vocabulary of operation types, such as clicking, scrolling, and pressing non-text keys, together with structured parameters. This output structure makes type-free GUI control a natural fit for Jev’s single-pass typed decisions~\cite{almeida2026jev,typesafe2026systemone}.
Existing GUI-oriented Jev interfaces select targets indexed by the Document Object Model (DOM) or supplied form-element options~\cite{browseruse2026jevultrafast,cuaai2026cuas1forms}, which external controllers resolve into executable targets.
Such structured target information is not consistently available across desktop applications, whereas screenshots provide a common observation interface.
This motivates a screenshot-based approach that predicts grounded operations and coordinates directly, without relying on DOM indices or supplied form options.

To address this gap, we introduce \method, a GUI-specific multimodal Jev architecture for single-pass action selection and precise coordinate prediction. 
Specifically, GUI-JEV employs a shared visual pointer to select target patches from screenshot features, which are then refined by an $8\times8$ within-patch offset readout to obtain their coordinates.
Specialized heads predict action types and finite parameters, while four action slots and a length head produce an ordered block in one multimodal backbone pass~\cite{qwen2026qwen35}.
The policy predicts grounded actions from backbone states without generating intermediate reasoning text, sharing computation across up to four actions.

To support reliable \emph{type-free} control, we develop a \emph{replay-verified data flywheel} that provides reproducible demonstrations of navigation, selection, and manipulation.
Controlled task construction aligns instructions, initial states, and evaluation conditions, while teacher and supervisor roles guide trajectory collection.
Native task evaluation checks completion, and exact replay on a fresh desktop verifies that the recorded actions reproduce the outcome.
These demonstrations supply action-block and coordinate supervision within a training curriculum that combines visual grounding, general interaction, and \emph{type-free} strategy learning.

We further introduce \textbf{OSWorld-JEV}, a benchmark to evaluate grounded action selection in various \emph{type-free} workflows.
\textbf{OSWorld-JEV} comprises 129 tasks spanning six categories, \emph{e.g.}, shopping, file and information management, office editing, visual editing, configuration and settings, and business workflows.
Each task is validated through GUI execution to confirm that it can be completed entirely without \texttt{TYPE} action.
Tasks require interpreting existing content and respecting operational constraints across multi-step workflows.
The benchmark evaluates how effectively and responsively GUI agents complete such workflows.

As summarized in \cref{fig:efficiency}, \method leads the compared models on ScreenSpot-v2~\cite{wu2025atlas} and ScreenSpot-Pro~\cite{li2025screenspotpro}, scoring 97.72 and 79.95.
These scores exceed Qwen3.8-Max by 4.95 and 6.52 points, respectively.
On OSWorld-JEV, it reaches 48.84 at 1.65\,s per step, exceeding Qwen-CUA by 5.44 points with a $4.30\times$ relative speedup.
These results position \method as a low-latency policy for various \emph{type-free} tasks, combining precise localization with responsive action selection.
\method separates action selection from text generation for general desktop tasks.
The policy determines every action and its execution order, while an on-demand writer supplies text payloads for \texttt{TYPE} actions.
This composition extends the same action-selection architecture to general desktop tasks, reaching 62.33 on OSWorld-Verified~\cite{xie2024osworld}.

In summary, our contributions are threefold:
\begin{itemize}
    \item \textbf{Precise, low-latency grounded control.} GUI-JEV predicts screen coordinates and ordered action blocks in one backbone pass, leading evaluated models on ScreenSpot-v2 and ScreenSpot-Pro and improving OSWorld-JEV completion over Qwen-CUA by 5.44 points at $4.30\times$ relative speed.
    \item \textbf{Replay-verified supervision.} Our data flywheel uses native evaluation and exact replay on fresh desktops to verify task completion and action-sequence reproducibility for \emph{type-free} learning.
    \item \textbf{Execution-verified evaluation.} OSWorld-JEV tests grounded action selection and constraint-aware execution across 129 tasks in six everyday \emph{type-free} categories.
\end{itemize}

%% file: fig/architecture_float.tex
\begin{figure}[!t]
    \centering
    \includegraphics[width=\textwidth]{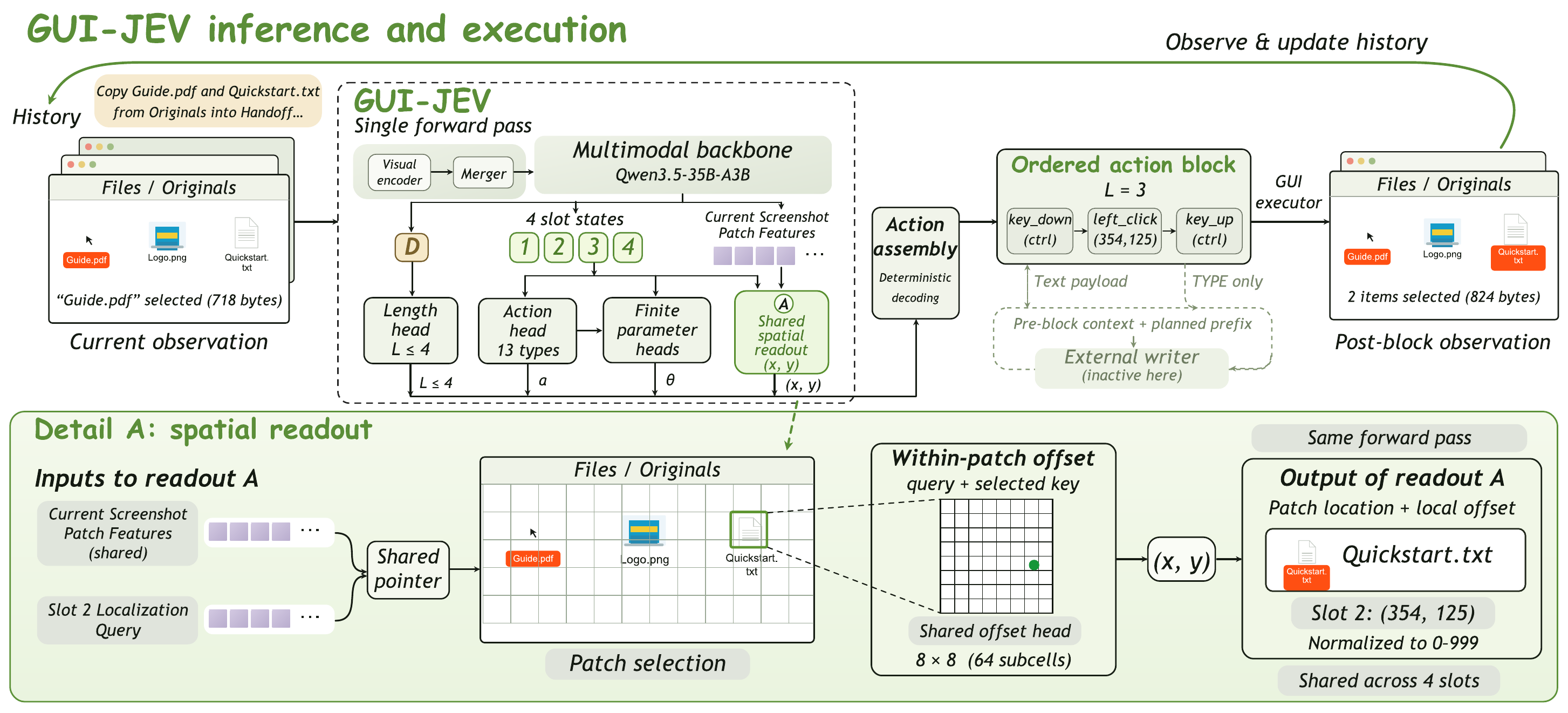}
    \caption{\textbf{Slot-based structured prediction for GUI control.}
    A slot-based interface produces ordered action blocks in a single multimodal backbone pass.
    Shared heads predict operations and finite parameters, with a length head selecting the active slots.
    A shared visual pointer and $8\times8$ within-patch offset head equip each slot with native screen coordinates.
    Desktop execution and observation updates close the interaction loop.}
    \label{fig:architecture}
\end{figure}

%% file: sec/2_related_work.tex
\section{Related work}
\label{sec:related_work}

\subsection{Computer-use agents}
SeeClick~\cite{cheng2024seeclick} and OS-ATLAS~\cite{wu2025atlas} advance screenshot grounding, while Aguvis~\cite{xu2024aguvis} and UI-TARS~\cite{qin2025uitars} combine grounding with planning.
GUI-Actor~\cite{wu2025guiactor} aligns action tokens with visual patches, informing our spatial readout.
GUI-Owl-1.5~\cite{xu2026guiowl} and Qwen-CUA~\cite{lu2026qwencua} extend control across applications and platforms.
\method enables responsive \emph{type-free} control through single-pass action and coordinate prediction, separating decision-making from text generation.
CUA-Gym~\cite{wang2026cuagym}, SCALECUA~\cite{lv2026scalecua}, and CUA-Master~\cite{yu2026cuamaster} scale supervision through task construction, verifiable synthesis, and feedback-driven collection.
Our flywheel verifies \emph{type-free} demonstrations via native scoring and fresh-desktop replay.
Complementing OSWorld~\cite{xie2024osworld}, OSWorld-JEV evaluates \emph{type-free} decisions across six workflow categories.

\subsection{Jev and structured decision models}
Jev predicts typed decisions over supplied alternatives~\cite{almeida2026jev,typesafe2026systemone}.
Duan et al.~\cite{duan2026general} introduce JEVal, study multi-step composition, and distill reasoning into InnerJev's single-pass readouts.
Visual Jev~\cite{yu2026visual} shares visual context across questions while retaining the language-model head; VideoJev~\cite{zhu2026videojev} scores temporal intervals and textual options.
Jev Ultrafast~\cite{browseruse2026jevultrafast} selects DOM-indexed targets, while CUA-S1 \texttt{cua-s1-forms}~\cite{cuaai2026cuas1forms} independently scores textual form-element options; both rely on external controllers to resolve executable targets.
\method combines a shared visual pointer, within-patch offsets, and fixed-schema heads to predict precise coordinates and ordered actions in one backbone pass.
On OSWorld-Verified, the policy composes with an on-demand writer for general desktop tasks.

%% file: sec/3_methodology.tex
\section{Method}
\label{sec:method}

\method predicts an ordered GUI action block in one multimodal forward pass, as shown in \cref{fig:architecture}.
\Cref{fig:structured_readout} details two structured readouts over the backbone states: action and parameter selection, and screenshot-to-coordinate grounding.

\subsection{Structured action prediction}
\input{fig/structured_readout_float}
At decision step $t$, four learned action-slot markers and one length marker are appended to the task instruction, execution history, and current screenshot $I_t$.
The Qwen3.5-35B-A3B backbone~\cite{qwen2026qwen35} produces slot states $q_1,\ldots,q_4$, a length state $d$, and current-image token states $K=\{k_1,\ldots,k_M\}$.
All slots share the same pre-block evidence.

In the action-block readout of \cref{fig:structured_readout}, a shared projection maps the marker states to the readout space.
The length head predicts $L_t\in\{1,2,3,4\}$ from the projected state of $d$, activating the first $L_t$ slots.
For each slot, a shared action head predicts one of 13 operation types.
The predicted action embedding is added to the projected slot state before shared finite-parameter heads select applicable fields, such as mouse buttons, key combinations, and scroll amounts.
Key combinations are chosen as full codebook entries.

For spatial actions, each slot also queries the coordinate readout below.
A deterministic compiler converts active slots into tool calls, which execute in order before a new screenshot and execution receipts update the context.

\WFclear
\subsection{Visual grounding}
The native coordinate readout in \cref{fig:structured_readout} combines patch selection with within-patch refinement.
Following GUI-Actor's action-to-patch alignment~\cite{wu2025guiactor}, a shared pointer queries current-image states using each slot state $q_i$.
Historical images inform the context but do not supply coordinate candidates.
With separately normalized and projected states $\bar q_i,\bar k_j$ of width $d_p=512$, the pointer computes
\begin{equation}
    p_{ij} =
    \operatorname{softmax}_{j}
    \left[\frac{\bar q_i^{\mathsf T}\bar k_j}{\sqrt{d_p}}\right],
    \qquad
    j_i^\star = \arg\max_j p_{ij}.
    \label{eq:pointer}
\end{equation}
The shared offset head takes the unprojected states $[q_i;k_{j_i^\star}]$ and classifies one of $8\times8$ subcells within the selected patch.
For patch row $r_i$, column $c_i$, and subcell indices $u_i,v_i\in\{0,\ldots,7\}$ in a $G_h\times G_w$ image-token grid, the normalized target is
\begin{equation}
    \tilde x_i = \frac{8c_i + u_i + 1/2}{8G_w},
    \quad
    \tilde y_i = \frac{8r_i + v_i + 1/2}{8G_h}.
    \label{eq:coordinate}
\end{equation}
The grid follows the processed image size. Coordinates are scaled by 1,000, rounded, and clipped to $[0,999]$ for execution.
Pointer and offset weights are shared across all slots.

\subsection{Three-stage training}
\method progressively learns visual localization, grounded action blocks, and \emph{type-free} strategies, carrying the same output architecture across all three stages.

\textbf{Stage 1: Visual grounding.}
We begin by training the first slot's shared pointer and offset heads on screenshot-target pairs.
The pointer learns a normalized Gaussian target $\pi_j$ ($\sigma=1$) over the annotated patch and its immediate neighbors, while the offset head learns the 64-way subcell label $z^{\mathrm{gt}}$.
For target patch $j^{\mathrm{gt}}$ and offset distribution $o_1$, the losses are
\begin{equation}
\begin{aligned}
    \mathcal{L}_{\mathrm{pointer}} &= -\sum_{j=1}^{M}\pi_j\log p_{1j},\\
    \mathcal{L}_{\mathrm{offset}} &= -\log o_1(z^{\mathrm{gt}}\mid j^{\mathrm{gt}}).
\end{aligned}
\label{eq:grounding_loss}
\end{equation}
The two terms supervise patch selection and within-patch localization, respectively.
We use the target patch for offset training and the predicted patch at inference.
Non-spatial losses are masked for grounding examples.

\textbf{Stage 2: General GUI control.}
We extend this localization capability to ordered action blocks using GUI demonstrations.
$\mathcal{L}_{\mathrm{length}}$ and $\mathcal{L}_{\mathrm{action}}$ supervise the active-slot count and per-slot operation types.
Demonstrated actions occupy consecutive slots in execution order, so the labels supervise both action selection and block ordering.
All slots are supervised from the same pre-block context, without intermediate screenshots.
$\mathcal{L}_{\mathrm{keys}}$ supervises complete key-combination entries, while $\mathcal{L}_{\mathrm{finite}}$ aggregates the remaining finite fields $\mathcal{F}$, such as mouse button, click count, and scroll amount.
For an interaction example, the cross-entropy losses are
\begin{equation}
\begin{aligned}
    \mathcal{L}_{\mathrm{length}} &= -\log p_{\mathrm{len}}(L^{\mathrm{gt}}\mid d),\\
    \mathcal{L}_{\mathrm{action}} &= -\bigl\langle\log p_{\mathrm{act}}(a_i^{\mathrm{gt}}\mid q_i)\bigr\rangle_{\mathrm{act}},\\
    \mathcal{L}_{\mathrm{keys}} &= -\bigl\langle\log p_{\mathrm{keys}}(y_{i,\mathrm{keys}}^{\mathrm{gt}}\mid q_i,a_i^{\mathrm{gt}})\bigr\rangle_{\mathrm{keys}},\\
    \mathcal{L}_{\mathrm{finite}} &= -\sum_{f\in\mathcal{F}}\bigl\langle\log p_f(y_{if}^{\mathrm{gt}}\mid q_i,a_i^{\mathrm{gt}})\bigr\rangle_f.
\end{aligned}
\label{eq:action_losses}
\end{equation}
Here $p$ denotes the corresponding head distribution, and $L^{\mathrm{gt}}$, $a_i^{\mathrm{gt}}$, and $y_{if}^{\mathrm{gt}}$ are the labeled length, operation, and parameter value.
$\langle\cdot\rangle_h$ denotes per-head masked slot averaging with the label weights described below, returning zero when no labels apply.
During training, parameter heads receive the labeled action type without exposing it to the backbone; at inference, they use the predicted type.
Pointer and offset losses apply to all labeled spatial slots in interaction examples.
We also mix in Stage 1 grounding examples to retain localization accuracy.

\textbf{Stage 3: Type-free adaptation.}
We adapt the model to type-free tasks using replay-verified demonstrations from \cref{sec:data_flywheel}.
Stage 3 retains the full 13-operation vocabulary, including \texttt{TYPE}, and the cross-entropy objective while using type-free instructions to supervise navigation, selection, and manipulation of existing content.
We retain general-interaction and grounding examples from earlier stages to preserve general control and localization skills.
The objective, parameter codebooks, and coordinate readouts remain unchanged.

All stages use the same output architecture and a unified structured objective with sample-dependent supervision masks:
\begin{equation}
\begin{aligned}
    \mathcal{L}_{\mathrm{GUI}} ={}&
    \mathcal{L}_{\mathrm{length}} + \mathcal{L}_{\mathrm{action}}
    + \mathcal{L}_{\mathrm{keys}} \\
    &+ 0.5\,\mathcal{L}_{\mathrm{finite}}
    + \mathcal{L}_{\mathrm{pointer}} + \mathcal{L}_{\mathrm{offset}}.
\end{aligned}
\label{eq:training}
\end{equation}
Each head loss is masked by label availability and field applicability, with per-label weights applied before averaging over supervised slots; block length has one label per example.
Grounding examples, including replay, activate only the pointer and offset terms; interaction examples activate all terms applicable to their labeled outputs.
Sample weights apply to the summed loss. Language-model labels remain masked.
Data sources and training configurations are given in \cref{sec:training_pipeline}.

%% file: fig/structured_readout_float.tex
\begin{wrapfigure}{R}{0.48\textwidth}
    \vspace{-\intextsep}
    \centering
    \includegraphics[width=\linewidth]{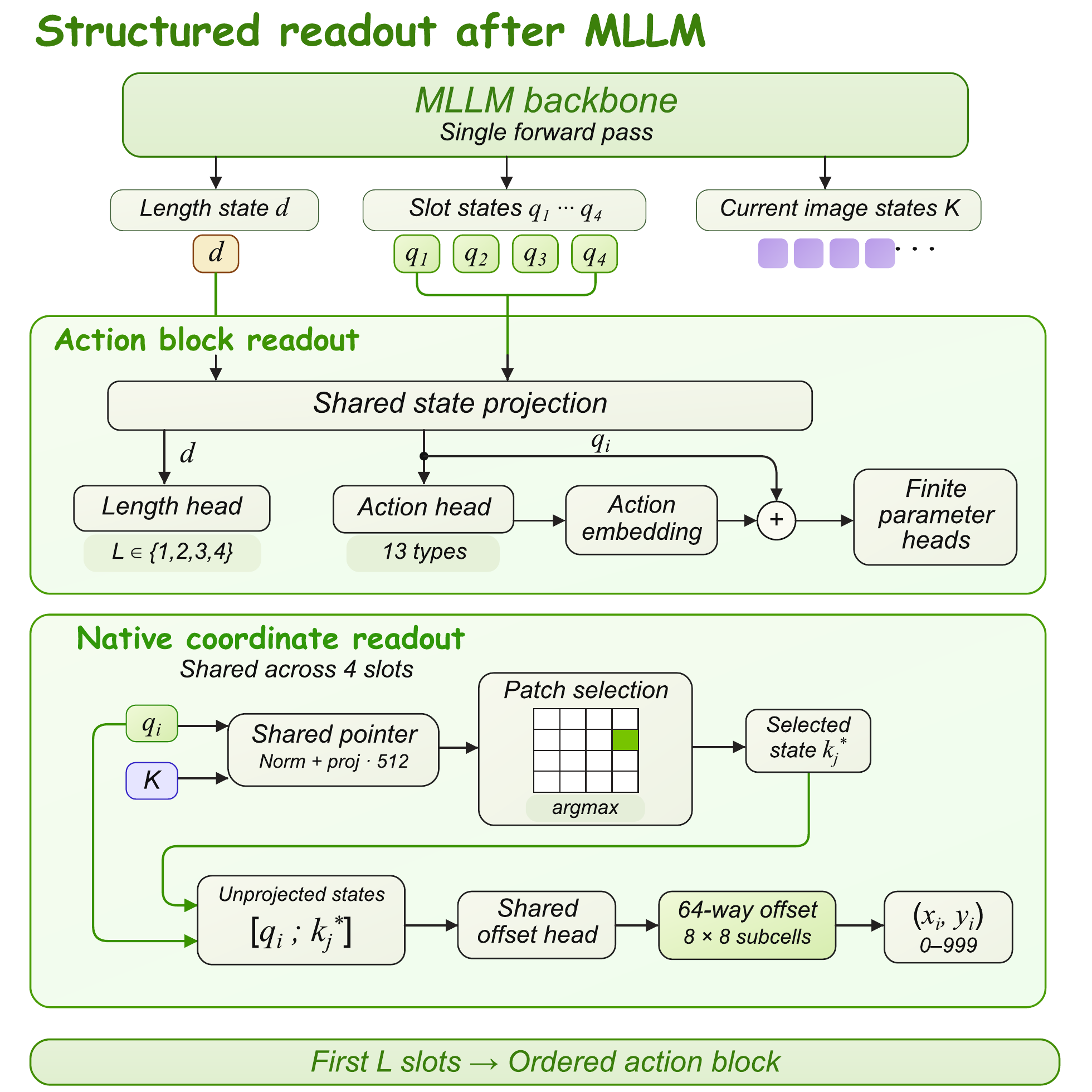}
    \caption{\textbf{Structured readout from shared backbone states.}
    Action-block heads predict length, operations, and action-conditioned parameters. Shared pointer and offset heads ground each slot in the current screenshot.}
    \label{fig:structured_readout}
\end{wrapfigure}

%% file: sec/data_flywheel.tex
\input{fig/data_flywheel_float}
\section{Type-free data flywheel}
\label{sec:data_flywheel}

The flywheel in \cref{fig:data_flywheel} collects reproducible type-free demonstrations through controlled task construction, supervised GUI rollouts, and replay verification.
A rule-based generator varies seed-task assets, layouts, and operational conditions while keeping instructions, initial states, target conditions, and preservation requirements consistent.
Structural checks, instruction-evaluator consistency checks, and semantic deduplication filter candidates before DSV4.1~\cite{deepseek2026v41} selects workflows.

Separate DSV4.1 teacher and supervisor calls guide collection: the teacher acts from public instructions, screenshots, and history, while the supervisor checks requirement coverage and proposes corrections.
Collection permits native GUI operations, including copying and pasting existing content, but prohibits TYPE, external clipboard injection, and terminal or script-based solving.
The executor records validated actions, receipts, and observations.
Acceptance requires successful termination, a passing native score, and a complete trajectory audit, followed by exact action replay on a fresh desktop without replanning and a second passing score.
Deterministic conversion maps actual pre-block screenshots, public instructions, history, and executed actions to structured supervision, preserving the action schema and loss masks.
Teacher reasoning, supervisor advice, and private evaluator information are excluded from policy supervision.
Image-reference, context-length, and target-index checks validate each batch.
Outcome and coverage feedback guide DSV4.1's subsequent task selection, permitted recipe changes, and recollection.

%% file: fig/data_flywheel_float.tex
\begin{figure}[!t]
    \centering
    \includegraphics[width=\textwidth]{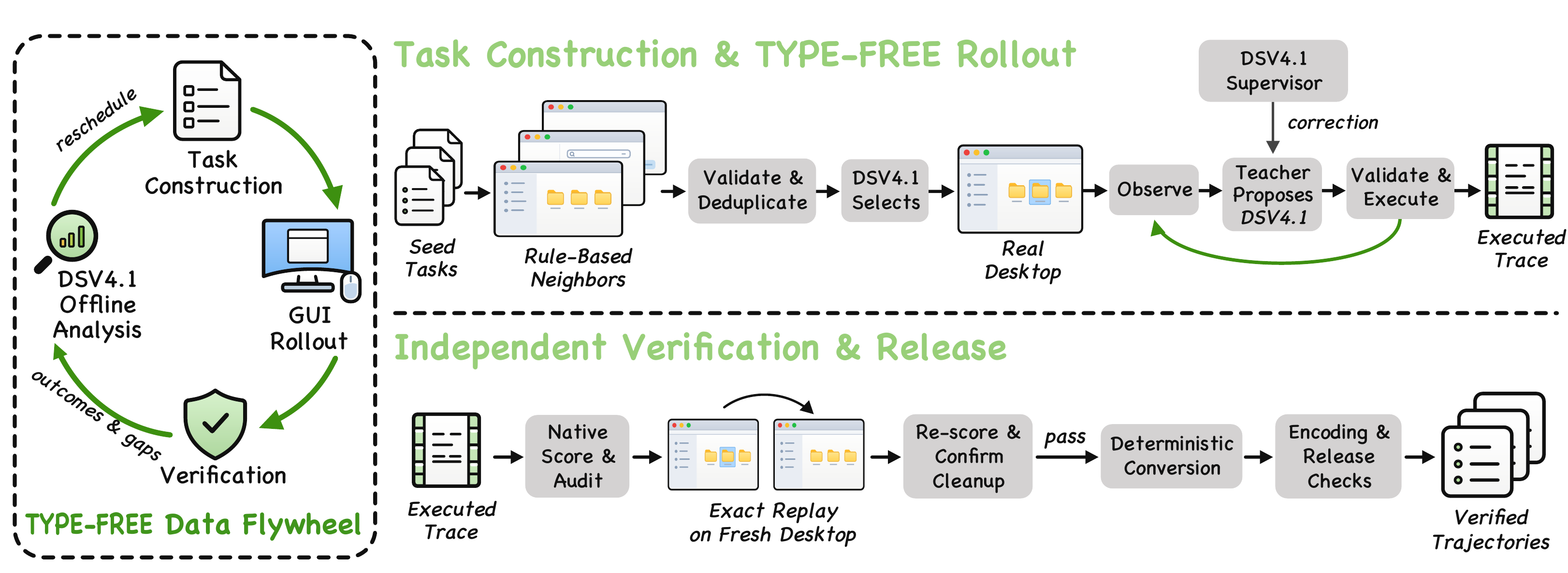}
    \caption{\textbf{Replay-verified supervision for type-free strategy learning.}
    Controlled tasks feed supervised desktop collection. Native scoring and exact replay on fresh desktops verify trajectories before conversion into structured supervision.
    Outcome and coverage feedback guide subsequent collection.}
    \label{fig:data_flywheel}
\end{figure}

%% file: sec/benchmark.tex
\input{fig/benchmark_float}
\section{OSWorld-JEV benchmark}
\label{sec:benchmark}

OSWorld-JEV evaluates grounded action selection and constraint-aware execution on everyday workflows with execution-verified type-free solutions.
It complements general computer-use evaluation~\cite{xie2024osworld} by testing operation selection, precise targeting, and constraint tracking across changing screens, together with interaction responsiveness.

\subsection{Task coverage}
The 129 tasks span six categories, as shown in \cref{fig:benchmark}.
\emph{Shopping} includes 18 tasks for constraint-based comparison and selection.
\emph{File and Information Management} includes 28 tasks for organizing files and correspondence.
\emph{Office Editing} includes 23 tasks for selective spreadsheet, document, and presentation changes.
\emph{Visual Editing} includes 12 tasks for image and video manipulation.
\emph{Configuration and Settings} includes 33 tasks for persistent preference and permission changes.
\emph{Business Workflows} includes 15 tasks for rule-governed record updates in GLPI, Odoo, and Planka.
Success requires making the requested changes while preserving unrelated content and protected state.

\subsection{Task construction and validation}
Every OSWorld-JEV task has an execution-verified solution using type-free GUI actions.
Tasks are adapted from existing open-source projects and workflows.
Each task pairs a public instruction with prepared assets, an initialized application state, and a native evaluator that checks the resulting state or saved artifacts.
Instructions specify target changes and preservation, saving, or final-screen requirements. For example, spreadsheet edits must leave untargeted rows intact.
Infrastructure handles initialization and private evaluation, while agents observe screenshots and execute GUI actions.
Task identities, assets, and evaluator versions support repeatable evaluation.
Task adaptation uses GPT-6 Astra~\cite{openai_gpt6_astra}, and solution validation uses GUI execution through \texttt{codex exec}.

%% file: fig/benchmark_float.tex
\begin{figure}[!t]
    \centering
    \includegraphics[width=\textwidth]{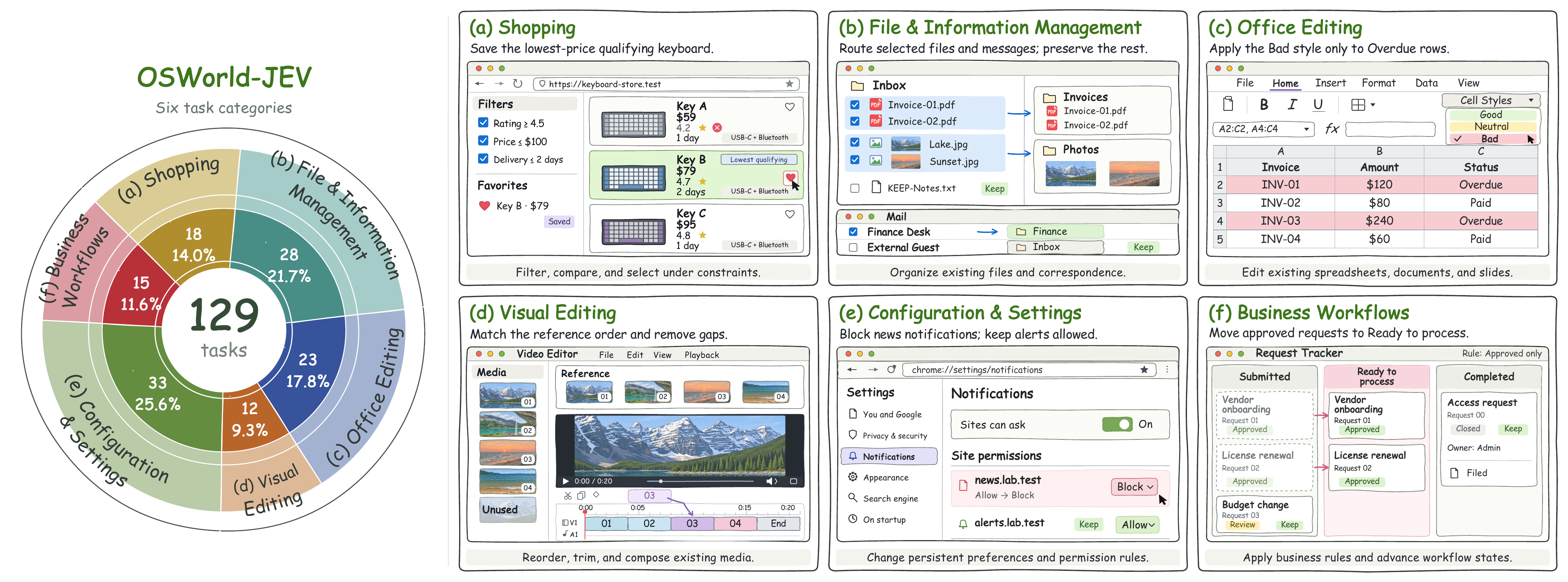}
    \caption{\textbf{OSWorld-JEV covers 129 tasks across six type-free categories.}
    The distribution and schematic examples highlight constraint-based selection, selective modification, and preservation of unrelated content.
    Tasks span shopping, file and information management, office editing, visual editing, configuration and settings, and business workflows.}
    \label{fig:benchmark}
\end{figure}

%% file: sec/4_experiments.tex
\section{Experiments}
\label{sec:experiments}
\subsection{Training pipeline}
\label{sec:training_pipeline}
\method uses the Qwen3.5-35B-A3B backbone and structured readouts described in \cref{sec:method}.
The visual encoder remains frozen. After a head-and-marker warmup, joint updates train the language backbone, visual merger, and relevant readout heads.
Stage 1 trains exclusively on grounding data from ShowUI-Web~\cite{lin2025showui}, GroundCUA-Train~\cite{feizi2026grounding}, the desktop and mobile subsets of OS-ATLAS~\cite{wu2025atlas}, JEDI Icons~\cite{xie2026scaling}, and ShowUI-Desktop~\cite{lin2025showui}.
After processing, these sources provide approximately 2 million training examples for screenshot-based target localization through the shared pointer and offset heads.
Stage 2 uses ProCUA data~\cite{jung2026procua} for general GUI interaction, together with grounding replay from Stage 1.
Stage 3 combines approximately 100,000 replay-verified OSWorld-JEV type-free training examples from \cref{sec:data_flywheel} with Stage 2 replay data and Stage 1 grounding replay.

\subsection{Benchmarks}
We evaluate five benchmarks spanning target localization and interactive task completion.
ScreenSpot-v2~\cite{wu2025atlas}, ScreenSpot-Pro~\cite{li2025screenspotpro}, and OSWorld-G~\cite{xie2026scaling} test the precision of screenshot-based grounding.
OSWorld-Verified~\cite{xie2024osworld} evaluates general desktop interaction.
For \method, selecting \texttt{TYPE} invokes GPT-5.6 Sol~\cite{openai_gpt56_sol} as an on-demand writer for the text payload, while the policy determines all actions and their execution order.
The writer receives the pre-block context and preceding planned actions, explicitly marked as not yet executed.
OSWorld-JEV contains 129 tasks designed for type-free completion across the six categories defined in \cref{sec:benchmark}.
These benchmarks connect spatial accuracy to completed workflows under both execution modes.

\subsection{Evaluation}
We evaluate grounding precision and interaction responsiveness in \cref{tab:main_results}, identify workflow-specific capabilities in \cref{tab:jev_categories}, and examine gains across training stages in \cref{tab:stage_ablation}.
Comparisons include the Qwen3.5 and Qwen3.8 families~\cite{qwen2026qwen35,qwen2026qwen38,qwen38_27b}, GUI-Owl-1.5~\cite{xu2026guiowl}, MAI-Gym~\cite{tong2026maigym}, Qwen-CUA~\cite{lu2026qwencua}, and structured decision models.
VideoJev~\cite{zhu2026videojev}, Visual Jev~\cite{yu2026visual}, and SII-DECISION~\cite{sii2026decision} are evaluated on grounding tasks; CUA-S1~\cite{cuaai2026cuas1fourb02} and \method additionally undergo interactive evaluation on OSWorld-Verified and OSWorld-JEV.
\method retains pre-action screenshots, ordered calls, and execution receipts for up to four recent events, summarizing earlier events as action-and-result text.
On OSWorld-JEV, all non-Jev baselines retain \texttt{TYPE} access, whereas \method masks \texttt{TYPE} and disables external writers while allowing keyboard shortcuts and GUI-based copying and pasting of existing content.
Completion is assessed by the benchmark's native evaluator.
Qwen3.8-Max is accessed via API and excluded from plotted score--time comparisons.

%% file: sec/5_results_and_discussions.tex
\input{sec/table_main_results}
\section{Results and discussion}
\label{sec:results}

\subsection{Overall benchmark results}
As shown in \cref{tab:main_results}, \method leads the compared models on ScreenSpot-v2 and ScreenSpot-Pro with 97.72 and 79.95, outperforming Qwen3.8-Max (2.4T) by 4.95 and 6.52 points.
It exceeds Qwen3.8-27B by 1.49 and 5.06 points at $4.65\times$ and $5.96\times$ relative speed.
On OSWorld-G, it reaches 75.58, surpassing GUI-Owl-1.5-32B by 9.86 points at $10.27\times$ relative speed.
It records the lowest reported timing on ScreenSpot-Pro and OSWorld-G, and outperforms the strongest grounding-only structured baseline, SII-DECISION-27B, by 46.30, 69.39, and 50.93 points across the three benchmarks.

On interactive benchmarks, \method achieves 62.33 with an on-demand writer on OSWorld-Verified and 48.84 without \texttt{TYPE} on OSWorld-JEV, with the lowest reported timing on both.
Against Qwen-CUA (397B-A17B), it improves OSWorld-JEV completion by 5.44 points at $4.30\times$ relative speed.
On OSWorld-Verified, \method delivers $2.72\times$ relative speed over Qwen-CUA, with completion scores of 62.33 and 63.40, respectively.
It outperforms GUI-Owl-1.5-32B by 9.42 and 39.54 points at $1.79\times$ and $3.25\times$ relative speed, and the Qwen3.5-35B base model with the same 35B-A3B backbone by 23.23 and 29.44 points, respectively.
These results support separating grounded action selection from text generation for responsive GUI control.

\input{sec/table_jev_categories}
\subsection{OSWorld-JEV results}
\method completes 48.84\% of OSWorld-JEV tasks through type-free execution with no external writer calls. Its 1.65\,s per step is the lowest reported timing on this benchmark (\cref{tab:main_results,tab:jev_categories}).
Against the 397B-parameter Qwen3.5 model, \method raises completion by 20.14 points at $4.84\times$ relative speed.
For latency-sensitive interaction, \method provides a $5.32\times$ relative speedup over the dense Qwen3.8-27B under the reported timing metric; their completion scores are 48.84 and 53.50, respectively.

Constraint-based shopping is a standout capability: \method achieves the highest completion rate among evaluated models at 77.78\%, exceeding the 2.4T-parameter Qwen3.8-Max's 72.2\% with a 35B-total backbone, approximately $69\times$ smaller in total parameter count.
It also reaches 67.86 in file and information management, 51.52 in configuration and settings, and 53.33 in business workflows, improving over its same-size base model by 17.86, 39.42, and 33.33 points, respectively.
These results highlight constraint-based shopping and extend the gains over the same-backbone baseline to file management, configuration, and business workflows.

\input{sec/table_stage_ablation}
\subsection{Grounding and general-control gains after adaptation}
The final mixed-data adaptation stage improves all four previously evaluated benchmarks (\cref{tab:stage_ablation}). It adds 0.16 points on ScreenSpot-v2, 4.18 on ScreenSpot-Pro, 3.42 on OSWorld-G, and 1.11 on OSWorld-Verified, while reaching 48.84 on OSWorld-JEV.
This stage combines replay-verified type-free demonstrations with general-interaction and grounding replay, coupling type-free task learning with gains in spatial precision and writer-assisted general control.
General GUI training also strengthens the grounding foundation: Stage 2 improves over Stage 1 by 2.91, 18.53, and 3.90 points on the three grounding benchmarks and reaches 61.22 on OSWorld-Verified.

%% file: sec/table_main_results.tex
\begin{table}[!t]
    \centering
    \caption{\textbf{Overall results across five GUI benchmarks.}
    Score reports each benchmark's overall result, with higher values indicating better performance.
    Time reports the corresponding benchmark-specific timing statistic, with lower values indicating faster execution.
    GUI-JEV's OSWorld-JEV time is measured in seconds per step.
    On OSWorld-JEV, all non-Jev baselines are allowed TYPE, while GUI-JEV uses type-free execution.
    A dash indicates an unavailable measurement or an inapplicable evaluation.
    Qwen3.8-Max is accessed via API and excluded from plotted comparisons of score and time.
    Empty cells indicate unreported measurements.}
    \label{tab:main_results}
    \fontsize{8}{9.5}\selectfont
    \setlength{\tabcolsep}{3pt}
    \resizebox{\linewidth}{!}{%
    \begin{tabular}{l|c|cc|cc|cc|cc|cc}
        \toprule
        \multirow{2}{*}{\textbf{Method}} & \multirow{2}{*}{\textbf{Size}}
        & \multicolumn{2}{c|}{\textbf{ScreenSpot-v2}~\cite{wu2025atlas}}
        & \multicolumn{2}{c|}{\textbf{ScreenSpot-Pro}~\cite{li2025screenspotpro}}
        & \multicolumn{2}{c|}{\textbf{OSWorld-G}~\cite{xie2026scaling}}
        & \multicolumn{2}{c|}{\textbf{OSWorld-Verified}~\cite{xie2024osworld}}
        & \multicolumn{2}{c}{\textbf{OSWorld-JEV}} \\
        \cmidrule(lr){3-4}\cmidrule(lr){5-6}\cmidrule(lr){7-8}\cmidrule(lr){9-10}\cmidrule(lr){11-12}
        & & Score & Time & Score & Time & Score & Time & Score & Time & Score & Time \\
        \midrule
        \multicolumn{12}{l}{\textbf{Frontier models}} \\
        Qwen3.8-Max~\cite{qwen2026qwen38} & 2.4T & \cellcolor{gray!20}92.77 & 2.847 & \cellcolor{gray!20}73.43 & 4.08 & \cellcolor{gray!20}77.48 & 2.67 & \cellcolor{gray!20}77.84 & - & \cellcolor{gray!20}78.29 & - \\
        Qwen3.8-27B~\cite{qwen38_27b} & 27B & \cellcolor{gray!20}96.23 & 1.94 & \cellcolor{gray!20}74.89 & 2.93 & \cellcolor{gray!20}66.84 & 2.41 & \cellcolor{gray!20}60.11 & 26.86 & \cellcolor{gray!20}53.50 & 8.77 \\
        Qwen3.5-397B~\cite{qwen2026qwen35} & 397B & \cellcolor{gray!20}94.50 & 5.17 & \cellcolor{gray!20}69.26 & 25.89 & \cellcolor{gray!20}69.50 & 24.37 & \cellcolor{gray!20}49.31 & 7.13 & \cellcolor{gray!20}28.70 & 7.99 \\
        \midrule
        \multicolumn{12}{l}{\textbf{GUI agents}} \\
        GUI-Owl-1.5-32B~\cite{xu2026guiowl} & 32B & \cellcolor{gray!20}95.05 & 2.64 & \cellcolor{gray!20}71.28 & 2.91 & \cellcolor{gray!20}65.72 & 2.71 & \cellcolor{gray!20}52.91 & 4.67 & \cellcolor{gray!20}9.30 & 5.37 \\
        MAI-Gym~\cite{tong2026maigym} & 9B & \cellcolor{gray!20}89.47 & 0.58 & \cellcolor{gray!20}31.12 & 1 & \cellcolor{gray!20}43.62 & 0.53 & \cellcolor{gray!20}48.75 & 22.43 & \cellcolor{gray!20}0.80 & 9.21 \\
        Qwen-CUA~\cite{lu2026qwencua} & 397B-A17B & \cellcolor{gray!20}73.98 & 3.31 & \cellcolor{gray!20}42.19 & 3.24 & \cellcolor{gray!20}65.07 & 3.85 & \cellcolor{gray!20}63.40 & 7.1 & \cellcolor{gray!20}43.40 & 7.09 \\
        \midrule
        \multicolumn{12}{l}{\textbf{Structured decision models}} \\
        VideoJev~\cite{zhu2026videojev} & 35B-A3B & \cellcolor{gray!20}43.00 & 0.78 & \cellcolor{gray!20}8.03 & 0.912 & \cellcolor{gray!20}15.25 & 0.874 & - & - & - & - \\
        Visual Jev~\cite{yu2026visual} & 4B & \cellcolor{gray!20}18.63 & 1.33 & \cellcolor{gray!20}4.87 & 1.474 & \cellcolor{gray!20}12.41 & 1.225 & - & - & - & - \\
        SII-DECISION-4B~\cite{sii2026decision} & 4B & \cellcolor{gray!20}45.13 & 1.34 & \cellcolor{gray!20}8.35 & 1.346 & \cellcolor{gray!20}14.89 & 1.475 & - & - & - & - \\
        SII-DECISION-27B~\cite{sii2026decision} & 27B & \cellcolor{gray!20}51.42 & 2.41 & \cellcolor{gray!20}10.56 & 2.417 & \cellcolor{gray!20}24.65 & 2.453 & - & - & - & - \\
        CUA-S1~\cite{cuaai2026cuas1fourb02} & 4B & \cellcolor{gray!20}10.80 & 0.397 & \cellcolor{gray!20}10.00 & 0.879 & \cellcolor{gray!20}12.40 & 0.287 & \cellcolor{gray!20}1.00 & 5.04 & \cellcolor{gray!20}0.00 & 5.34 \\
        \midrule
        \shortstack[l]{Qwen3.5-35B~\cite{qwen2026qwen35}\\base model} & 35B-A3B & \cellcolor{gray!20}95.20 & 1.31 & \cellcolor{gray!20}67.40 & 8.98 & \cellcolor{gray!20}64.50 & 17.01 & \cellcolor{gray!20}39.10 & 11.57 & \cellcolor{gray!20}19.40 & 5.46 \\
        \rowcolor{resultgreen!50}
        \textbf{GUI-JEV} & \textbf{35B-A3B} & \textbf{97.72} & \textbf{0.417} & \textbf{79.95} & \textbf{0.492} & \textbf{75.58} & \textbf{0.264} & \textbf{62.33} & \textbf{2.61} & \textbf{48.84} & \textbf{1.65} \\
        \bottomrule
    \end{tabular}
    }
\end{table}

%% file: sec/table_jev_categories.tex
\begin{table}[!t]
    \centering
    \caption{\textbf{OSWorld-JEV results across six workflow categories.}
    Category headers give the number of tasks.
    All non-Jev baselines are allowed TYPE, while GUI-JEV uses type-free execution.
    Higher scores indicate better performance.
    Empty cells are reserved for results not reported or not evaluated.}
    \label{tab:jev_categories}
    \fontsize{8}{9.5}\selectfont
    \setlength{\tabcolsep}{4pt}
    \resizebox{\linewidth}{!}{%
    \begin{tabular}{l|c|cccccc|c}
        \toprule
        \multirow{3}{*}{\textbf{Method}} & \multirow{3}{*}{\textbf{Size}}
        & \textbf{Shopping} & \textbf{Visual} & \textbf{File and} & \textbf{Office} & \textbf{Configuration} & \textbf{Business} & \multirow{3}{*}{\textbf{Overall}} \\
        & & & \textbf{editing} & \textbf{information} & \textbf{editing} & \textbf{and settings} & \textbf{workflows} & \\
        & & 18 tasks & 12 tasks & 28 tasks & 23 tasks & 33 tasks & 15 tasks & \\
        \midrule
        \multicolumn{9}{l}{\textbf{Frontier models}} \\
        Qwen3.8-Max~\cite{qwen2026qwen38} & 2.4T & 72.2 & 8.3 & 89.3 & 82.6 & 90.9 & 86.7 & \cellcolor{gray!20}78.29 \\
        Qwen3.8-27B~\cite{qwen38_27b} & 27B & 0.0 & 8.3 & 89.3 & 52.2 & 63.6 & 66.7 & \cellcolor{gray!20}53.50 \\
        Qwen3.5-397B~\cite{qwen2026qwen35} & 397B & 5.6 & 0.0 & 67.9 & 21.7 & 21.2 & 33.3 & \cellcolor{gray!20}28.70 \\
        \midrule
        \multicolumn{9}{l}{\textbf{GUI agents}} \\
        GUI-Owl-1.5-32B~\cite{xu2026guiowl} & 32B & 0.0 & 8.3 & 17.9 & 0.0 & 12.1 & 13.3 & \cellcolor{gray!20}9.30 \\
        MAI-Gym~\cite{tong2026maigym} & 9B & 0.0 & 0.0 & 3.6 & 0.0 & 0.0 & 0.0 & \cellcolor{gray!20}0.80 \\
        Qwen-CUA~\cite{lu2026qwencua} & 397B-A17B & 11.1 & 25.0 & 71.4 & 52.2 & 33.3 & 53.3 & \cellcolor{gray!20}43.40 \\
        \midrule
        \multicolumn{9}{l}{\textbf{Structured decision models}} \\
        CUA-S1~\cite{cuaai2026cuas1fourb02} & 4B & 0.0 & 0.0 & 0.0 & 0.0 & 0.0 & 0.0 & \cellcolor{gray!20}0.00 \\
        \midrule
        \shortstack[l]{Qwen3.5-35B~\cite{qwen2026qwen35}\\base model} & 35B-A3B & 0.0 & 8.3 & 50.0 & 13.0 & 12.1 & 20.0 & \cellcolor{gray!20}19.40 \\
        \rowcolor{resultgreen!50}
        \textbf{GUI-JEV} & \textbf{35B-A3B} & 77.78 & 0 & 67.86 & 21.74 & 51.52 & 53.33 & \textbf{48.84} \\
        \bottomrule
    \end{tabular}
    }
\end{table}

%% file: sec/table_stage_ablation.tex
\begin{table}[!t]
    \centering
    \caption{\textbf{Results after each stage of the training curriculum.}
    Higher scores indicate better performance.
    Stage 3 is the final GUI-JEV checkpoint reported in \cref{tab:main_results}.
    Dashes indicate unreported evaluations.}
    \label{tab:stage_ablation}
    \fontsize{8}{9.5}\selectfont
    \setlength{\tabcolsep}{3pt}
    \resizebox{\linewidth}{!}{%
    \begin{tabular}{l|ccc|cc}
        \toprule
        \multirow{2}{*}{\textbf{Checkpoint}}
        & \multicolumn{3}{c|}{\textbf{Grounding}}
        & \multicolumn{2}{c}{\textbf{GUI interaction}} \\
        \cmidrule(lr){2-4}\cmidrule(lr){5-6}
        & ScreenSpot-v2 & ScreenSpot-Pro & OSWorld-G & OSWorld-Verified & OSWorld-JEV \\
        \midrule
        Stage 1: Grounding & 94.65 & 57.24 & 68.26 & - & - \\
        Stage 2: General agent with grounding replay & 97.56 & 75.77 & 72.16 & 61.22 & - \\
        \midrule
        \rowcolor{resultgreen!50}
        \textbf{Stage 3: type-free adaptation} & \textbf{97.72} & \textbf{79.95} & \textbf{75.58} & \textbf{62.33} & \textbf{48.84} \\
        \bottomrule
    \end{tabular}
    }
\end{table}

%% file: sec/6_conclusion.tex
\section{Conclusion}
\label{sec:conclusion}
GUI-JEV turns screenshot-based GUI control into single-pass structured action prediction, combining precise visual grounding with low-latency interaction without generating explicit reasoning traces.
It leads evaluated models on ScreenSpot-v2 and ScreenSpot-Pro and achieves 48.84\% completion at 1.65\,s per step on OSWorld-JEV---5.44 points higher completion at $4.30\times$ relative speed than Qwen-CUA.
Replay-verified supervision supports this policy, while an on-demand writer extends the same action-selection architecture to general desktop tasks.

%% file: sec/appendix.tex
\beginappendix
\phantomsection
\addcontentsline{toc}{section}{Appendix}
\setcounter{figure}{0}
\setcounter{table}{0}
\setcounter{equation}{0}
\renewcommand{\thefigure}{S\arabic{figure}}
\renewcommand{\thetable}{S\arabic{table}}
\renewcommand{\theequation}{S\arabic{equation}}
\renewcommand{\theHfigure}{appendix.\arabic{figure}}
\renewcommand{\theHtable}{appendix.\arabic{table}}
\renewcommand{\theHequation}{appendix.\arabic{equation}}
\begingroup
\setlength{\emergencystretch}{1.5em}
\newcommand{\code}[1]{\texttt{#1}}
\input{supplement/implementation}

\clearpage
\input{supplement/localization_precision}
\input{supplement/representative_trajectories}
\clearpage
\endgroup

%% file: supplement/implementation.tex
\section{Action Interface and Implementation Details}
\label{sec:supp-implementation}

\subsection{Complete operation vocabulary}
The policy predicts one of 13 operation classes in each of four action slots (\cref{tab:operations}), together with a block length, operation-specific finite parameters and spatial targets. In writer-enabled evaluation, an external writer supplies the text for a selected \code{TYPE} operation. OSWorld-JEV uses the type-free configuration: the \code{TYPE} logit is masked in every slot and the writer is disabled.

\begin{table}[!ht]
\centering\small
\caption{\textbf{All 13 operations and their parameters.} Square brackets denote optional fields. Coordinates are two-dimensional targets in $[0,999]^2$. Appendix~\ref{sec:supp-codebooks} lists the complete finite codebooks and their class IDs.}
\label{tab:operations}
\setlength{\tabcolsep}{4pt}
\begin{tabularx}{\textwidth}{@{}r l p{0.25\textwidth} X@{}}
\toprule ID & Operation & Active parameters & Deterministic execution semantics \\
\midrule
0 & \code{CLICK} & $(x,y)$, button, count & Left, right or middle click; count $\in\{1,2,3\}$. The native codec supports left single/double/triple and right/middle single clicks. \\
1 & \code{MOVE} & $(x,y)$ & Move the pointer to the target without clicking. \\
2 & \code{DRAG} & $(x,y)$, [duration] & Left-button drag from the current pointer position to the predicted target. \\
3 & \code{SCROLL} & axis, signed \code{pixels} & Scroll along the selected vertical or horizontal axis by the signed amount specified by one codebook entry. \\
4 & \code{KEY} & keys & Press one complete key or key-combination codebook entry. \\
5 & \code{KEY\_DOWN} & keys & Press and hold the selected key(s). \\
6 & \code{KEY\_UP} & keys & Release the selected key(s). \\
7 & \code{MOUSE\_DOWN} & button, [$(x,y)$] & Press the mouse button, optionally at a specified target. The native call codec supports the left-button form. \\
8 & \code{MOUSE\_UP} & button, [$(x,y)$] & Release the mouse button, optionally at a specified target. The native call codec supports the left-button form. \\
9 & \code{TYPE} & No policy parameter & Request a text payload from the separate writer, if the execution policy permits it. \\
10 & \code{WAIT} & time & Wait for the selected duration in seconds, subject to executor limits. \\
11 & \code{SCREENSHOT} & None & Request a screenshot observation. \\
12 & \code{TERMINATE} & status & End the attempt with a reported \code{success} or \code{failure} status. The native evaluator determines task success. \\
\bottomrule
\end{tabularx}
\end{table}

\paragraph{Finite parameters and validity.}
Button has three classes (left/right/middle), click count has three, axis has two (vertical/horizontal), and termination status has two. Coordinate presence and duration presence each have two classes. Keys, signed scroll amounts, waits and drag durations have 192, 82, 37 and 14 classes, respectively. Each key or ordered key combination is represented by one codebook entry; case and spelling aliases have separate class IDs. A binary presence prediction determines whether an optional duration is included. Coordinates are required for \code{CLICK}, \code{MOVE} and \code{DRAG}, and optional for \code{MOUSE\_DOWN} and \code{MOUSE\_UP}. The codec enforces the supported button/count combinations in \cref{tab:operations}.

\paragraph{Units.}
The scroll amount \code{pixels} is measured in signed wheel detents. Its magnitude gives the number of detents; negative values scroll down or left, and positive values scroll up or right, according to the selected axis. Wait and drag durations are measured in seconds. The finite vocabularies for these quantities are listed in Appendix~\ref{sec:supp-codebooks}.

\subsection{Four slots, length prediction and coordinates}
Let $q_1,\ldots,q_4$ be the backbone states at \code{gui\_slot\_0}, $\ldots$, \code{gui\_slot\_3}, and let $d$ be the state at \code{decide}. These five learned markers follow the public instruction, executed history and current screenshot. A shared LayerNorm--linear--GELU projection $f$ maps states to width 512. The length and operation predictions are
\begin{align}
 L &= 1+\arg\max_{\ell\in\{0,1,2,3\}} [W_Lf(d)]_\ell,\\
 a_i &= \arg\max_a [W_Af(q_i)]_a.
\end{align}
The predicted length activates slots $1,\ldots,L$. Each finite-parameter head reads $f(q_i)+e(a_i)$, where $e$ is a learned operation embedding. These heads are conditioned on the ground-truth operation during training and the predicted operation during inference.

\begin{figure}[!ht]
\centering
\includegraphics[width=\textwidth]{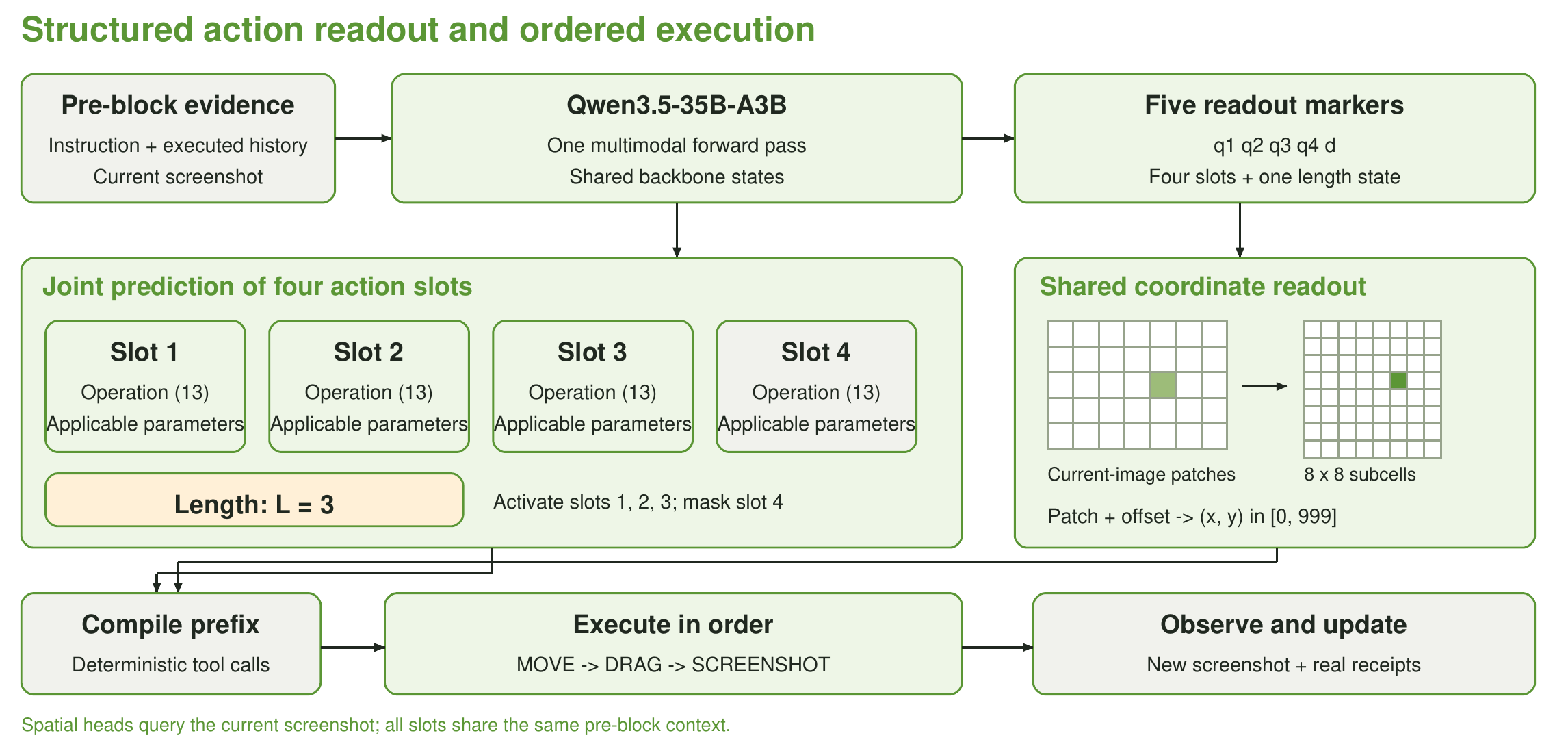}
\caption{\textbf{One observation, four ordered slots.} All slots share the instruction, history and screenshot available before the action block. The length readout activates a prefix, and shared spatial heads select an image token and a local subcell. The active slots are converted into executable calls and executed in order. The schematic illustrates a block of length $L=3$.}
\label{fig:supp-readout}
\end{figure}

\paragraph{Current-image candidates.}
The shared spatial pointer selects a target token from the current screenshot. Historical screenshots contribute context through the backbone. Separate LayerNorm and bias-free linear projections map each unprojected slot state and current-image token state to a 512-dimensional query and key, respectively. Their scaled dot products and softmax define
\begin{equation}
 p_{ij}=\operatorname{softmax}_j\!\left(\bar q_i^\top\bar k_j/\sqrt{512}\right),\qquad
 j_i^*=\arg\max_j p_{ij}.
\end{equation}
Pointer projection and dot products run in FP32. The offset head is LayerNorm--linear(512)--GELU--linear(64) over $[q_i;k_{j_i^*}]$. Its 64 classes index an $8\times8$ grid within the selected image token. For patch $(r,c)$ in the post-merge image-token grid $G_h\times G_w$ and subcell $(u,v)$,
\begin{equation}
\begin{aligned}
 x&=\operatorname{clip}_{[0,999]}\!\left(\operatorname{round}\!\left[1000\frac{c+(u+1/2)/8}{G_w}\right]\right),\\
 y&=\operatorname{clip}_{[0,999]}\!\left(\operatorname{round}\!\left[1000\frac{r+(v+1/2)/8}{G_h}\right]\right).
\end{aligned}
\end{equation}
The grid is obtained from the processed image dimensions with spatial merge size 2. Grounding supervision uses a Gaussian soft target over the target patch and its radius-1 neighborhood ($\sigma=1$). Offset training uses the target patch; inference uses the patch selected by the pointer.

\paragraph{Execution order and new observations.}
All active-slot calls are predicted jointly from the pre-block observation and executed sequentially. A \code{MOVE} can establish the starting point of a subsequent \code{DRAG}. A terminal call stops the attempt, and execution receipts, including errors, enter the action history. The next policy decision uses an updated screenshot and the accumulated execution history. The context contains up to four recent events with their pre-action images, the current image, and textual records of earlier actions and receipts. The model context limit is 32,768 tokens, with 26,624 tokens allocated to instructions, observations and history. Older event images are removed together as needed to meet this budget.

\paragraph{Objective.}
The structured loss is
\begin{equation}
\begin{split}
\mathcal L_{\mathrm{GUI}}={}&\mathcal L_L+\mathcal L_A+\mathcal L_{\mathrm{keys}}
 +0.5\mathcal L_{\mathrm{finite}}\\
 &+\mathcal L_{\mathrm{pointer}}+\mathcal L_{\mathrm{offset}}.
\end{split}
\end{equation}
Loss terms are masked according to slot activity and parameter applicability, and the summed structured loss is scaled by the sample weight. Grounding records supervise the pointer and offset heads. Training combines this structured objective with the MoE load-balancing loss, weighted by $0.001$. Type-free training uses 13-class operation cross-entropy; the \code{TYPE} mask is applied during evaluation.

\FloatBarrier
\subsection{Complete finite codebooks}
\label{sec:supp-codebooks}
The following tables list all finite parameter values and their zero-based class IDs, including aliases.
\input{supplement/generated/codebooks}

%% file: supplement/generated/codebooks.tex
{\fontsize{8}{9.2}\selectfont\setlength{\tabcolsep}{3pt}
\begin{longtable}{@{}r p{0.195\textwidth} r p{0.195\textwidth} r p{0.195\textwidth} r p{0.195\textwidth}@{}}
\caption{Complete keys codebook (192 entries). IDs are zero-based.}\label{tab:codebook-keys}\\
\toprule ID & Value & ID & Value & ID & Value & ID & Value \\ \midrule\endfirsthead
\toprule ID & Value & ID & Value & ID & Value & ID & Value \\ \midrule\endhead
0 & 1 & 48 & c & 96 & ctrl + shift + End & 144 & page\_down \\
1 & B & 49 & command & 97 & ctrl + shift + a & 145 & page\_up \\
2 & BackSpace & 50 & ctrl + , & 98 & ctrl + shift + b & 146 & pagedown \\
3 & D & 51 & ctrl + - & 99 & ctrl + shift + down & 147 & pageup \\
4 & Delete & 52 & ctrl + 0 & 100 & ctrl + shift + e & 148 & plus \\
5 & Down & 53 & ctrl + 1 & 101 & ctrl + shift + end & 149 & q \\
6 & E & 54 & ctrl + 2 & 102 & ctrl + shift + f9 & 150 & return \\
7 & End & 55 & ctrl + End & 103 & ctrl + shift + j & 151 & right \\
8 & Enter & 56 & ctrl + Home & 104 & ctrl + shift + n & 152 & s \\
9 & Escape & 57 & ctrl + ` & 105 & ctrl + shift + o & 153 & shift + Down \\
10 & F1 & 58 & ctrl + a & 106 & ctrl + shift + p & 154 & shift + End \\
11 & F10 & 59 & ctrl + alt + enter & 107 & ctrl + shift + right & 155 & shift + F8 \\
12 & F11 & 60 & ctrl + alt + p & 108 & ctrl + shift + s & 156 & shift + Page\_Up \\
13 & F2 & 61 & ctrl + alt + t & 109 & ctrl + shift + t & 157 & shift + Right \\
14 & F3 & 62 & ctrl + b & 110 & ctrl + shift + v & 158 & shift + alt + f \\
15 & F4 & 63 & ctrl + c & 111 & ctrl + shift + x & 159 & shift + ctrl + End \\
16 & F5 & 64 & ctrl + d & 112 & ctrl + t & 160 & shift + ctrl + Right \\
17 & F8 & 65 & ctrl + down & 113 & ctrl + u & 161 & shift + ctrl + a \\
18 & Home & 66 & ctrl + e & 114 & ctrl + v & 162 & shift + ctrl + down \\
19 & Left & 67 & ctrl + end & 115 & ctrl + w & 163 & shift + ctrl + e \\
20 & Next & 68 & ctrl + enter & 116 & ctrl + x & 164 & shift + ctrl + end \\
21 & PageDown & 69 & ctrl + equal & 117 & ctrl + z & 165 & shift + ctrl + f5 \\
22 & PageUp & 70 & ctrl + f & 118 & ctrl & 166 & shift + ctrl + j \\
23 & Page\_Down & 71 & ctrl + f10 & 119 & d & 167 & shift + ctrl + n \\
24 & Page\_Up & 72 & ctrl + f12 & 120 & delete & 168 & shift + down \\
25 & Prior & 73 & ctrl + f3 & 121 & down & 169 & shift + end \\
26 & R & 74 & ctrl + f9 & 122 & e & 170 & shift + f8 \\
27 & Return & 75 & ctrl + g & 123 & end & 171 & shift + o \\
28 & Right & 76 & ctrl + grave & 124 & enter & 172 & shift + pageup \\
29 & Tab & 77 & ctrl + h & 125 & esc & 173 & shift + right + right + right + right + right \\
30 & Up & 78 & ctrl + home & 126 & escape & 174 & shift + right \\
31 & a & 79 & ctrl + i & 127 & f & 175 & shift + tab \\
32 & alt + F12 & 80 & ctrl + j & 128 & f10 & 176 & shift \\
33 & alt + F4 & 81 & ctrl + k & 129 & f11 & 177 & space \\
34 & alt + Tab & 82 & ctrl + l & 130 & f12 & 178 & super + d \\
35 & alt + down & 83 & ctrl + m & 131 & f2 & 179 & super + down \\
36 & alt + enter & 84 & ctrl + minus & 132 & f3 & 180 & super + h \\
37 & alt + f & 85 & ctrl + n & 133 & f4 & 181 & super + right \\
38 & alt + f2 & 86 & ctrl + o & 134 & f5 & 182 & super + up \\
39 & alt + f4 & 87 & ctrl + p & 135 & f8 & 183 & super \\
40 & alt + i & 88 & ctrl + page + down & 136 & g & 184 & tab \\
41 & alt + left & 89 & ctrl + plus & 137 & grave & 185 & u \\
42 & alt + r & 90 & ctrl + q & 138 & h & 186 & up \\
43 & alt + tab & 91 & ctrl + r & 139 & home & 187 & v \\
44 & alt + z & 92 & ctrl + right & 140 & l & 188 & w \\
45 & alt & 93 & ctrl + s & 141 & left & 189 & win \\
46 & b & 94 & ctrl + shift + 5 & 142 & p & 190 & x \\
47 & backspace & 95 & ctrl + shift + Delete & 143 & page & 191 & z \\
\bottomrule\end{longtable}
\begin{longtable}{@{}r p{0.195\textwidth} r p{0.195\textwidth} r p{0.195\textwidth} r p{0.195\textwidth}@{}}
\caption{Complete pixels codebook (82 entries). IDs are zero-based.}\label{tab:codebook-pixels}\\
\toprule ID & Value & ID & Value & ID & Value & ID & Value \\ \midrule\endfirsthead
\toprule ID & Value & ID & Value & ID & Value & ID & Value \\ \midrule\endhead
0 & -1 & 21 & -30 & 42 & -900 & 63 & 300 \\
1 & -10 & 22 & -300 & 43 & 10 & 64 & 3000 \\
2 & -100 & 23 & -350 & 44 & 100 & 65 & 350 \\
3 & -1000 & 24 & -4 & 45 & 1000 & 66 & 4 \\
4 & -1100 & 25 & -40 & 46 & 10000 & 67 & 40 \\
5 & -12 & 26 & -400 & 47 & 120 & 68 & 400 \\
6 & -120 & 27 & -450 & 48 & 1200 & 69 & 450 \\
7 & -1200 & 28 & -5 & 49 & 1300 & 70 & 5 \\
8 & -15 & 29 & -50 & 50 & 15 & 71 & 50 \\
9 & -150 & 30 & -500 & 51 & 150 & 72 & 500 \\
10 & -1500 & 31 & -5000 & 52 & 1500 & 73 & 6 \\
11 & -180 & 32 & -507 & 53 & 1600 & 74 & 60 \\
12 & -2 & 33 & -550 & 54 & 2 & 75 & 600 \\
13 & -20 & 34 & -6 & 55 & 20 & 76 & 700 \\
14 & -200 & 35 & -60 & 56 & 200 & 77 & 8 \\
15 & -2000 & 36 & -600 & 57 & 2000 & 78 & 80 \\
16 & -20000 & 37 & -700 & 58 & 25 & 79 & 800 \\
17 & -25 & 38 & -750 & 59 & 250 & 80 & 8000 \\
18 & -250 & 39 & -8 & 60 & 2500 & 81 & 900 \\
19 & -2500 & 40 & -80 & 61 & 3 &  &  \\
20 & -3 & 41 & -800 & 62 & 30 &  &  \\
\bottomrule\end{longtable}
\begin{longtable}{@{}r p{0.195\textwidth} r p{0.195\textwidth} r p{0.195\textwidth} r p{0.195\textwidth}@{}}
\caption{Complete time codebook (37 entries). IDs are zero-based.}\label{tab:codebook-time}\\
\toprule ID & Value & ID & Value & ID & Value & ID & Value \\ \midrule\endfirsthead
\toprule ID & Value & ID & Value & ID & Value & ID & Value \\ \midrule\endhead
0 & 0.1 & 10 & 15 & 20 & 25 & 30 & 6 \\
1 & 0.3 & 11 & 150 & 21 & 3 & 31 & 60 \\
2 & 0.5 & 12 & 16 & 22 & 30 & 32 & 600 \\
3 & 0.8 & 13 & 18 & 23 & 300 & 33 & 7 \\
4 & 1 & 14 & 180 & 24 & 35 & 34 & 8 \\
5 & 1.2 & 15 & 2 & 25 & 4 & 35 & 9 \\
6 & 1.5 & 16 & 2.5 & 26 & 40 & 36 & 90 \\
7 & 10 & 17 & 20 & 27 & 45 &  &  \\
8 & 12 & 18 & 22 & 28 & 5 &  &  \\
9 & 120 & 19 & 240 & 29 & 50 &  &  \\
\bottomrule\end{longtable}
\begin{longtable}{@{}r p{0.195\textwidth} r p{0.195\textwidth} r p{0.195\textwidth} r p{0.195\textwidth}@{}}
\caption{Complete duration codebook (14 entries). IDs are zero-based.}\label{tab:codebook-duration}\\
\toprule ID & Value & ID & Value & ID & Value & ID & Value \\ \midrule\endfirsthead
\toprule ID & Value & ID & Value & ID & Value & ID & Value \\ \midrule\endhead
0 & 0.2 & 4 & 0.8 & 8 & 2 & 12 & 5 \\
1 & 0.3 & 5 & 1 & 9 & 2.5 & 13 & 8 \\
2 & 0.4 & 6 & 1.2 & 10 & 3 &  &  \\
3 & 0.5 & 7 & 1.5 & 11 & 4 &  &  \\
\bottomrule\end{longtable}
}

%% file: supplement/localization_precision.tex
\section{Within-patch localization precision}
\label{sec:supp-offset-ablation}

We isolate the contribution of within-patch precision through inference-time
quantization of a fixed checkpoint, holding model weights, image processing,
and patch predictions constant. The native offset prediction is quantized
to a $b\times b$ grid, where $b\in\{1,2,4,8\}$. The $1\times1$ setting returns
the patch center, and $8\times8$ retains the native prediction. We use the
native benchmark scorers and report scores as percentages. Confidence
intervals use paired task-level bootstrap with 10,000 resamples.

\begin{table}[!ht]
\centering
\caption{\textbf{Within-patch localization precision.} Accuracy improves as
the decoded coordinate grid becomes finer.}
\label{tab:supp-offset-ablation}
\small\setlength{\tabcolsep}{4pt}
\begin{tabular}{@{}lrrrr@{}}
\toprule
Benchmark & $1\!\times\!1$ & $2\!\times\!2$ & $4\!\times\!4$ & $8\!\times\!8$ \\
\midrule
ScreenSpot-v2  & 94.50 & 97.56 & 97.60 & \textbf{97.72} \\
ScreenSpot-Pro & 40.54 & 66.67 & 76.53 & \textbf{79.95} \\
OSWorld-G      & 50.89 & 68.26 & 73.58 & \textbf{75.58} \\
\bottomrule
\end{tabular}
\end{table}

Within-patch refinement improves grounding over the patch center by
3.22, 39.41, and 24.69 percentage points on ScreenSpot-v2, ScreenSpot-Pro,
and OSWorld-G, respectively, while keeping patch predictions fixed
(\cref{tab:supp-offset-ablation}). The largest gain occurs on ScreenSpot-Pro,
with a 95\% interval of $[36.94,41.87]$ points. Refining the grid from
$4\times4$ to $8\times8$ adds a further 3.42 points on ScreenSpot-Pro,
with an interval of $[2.47,4.43]$.

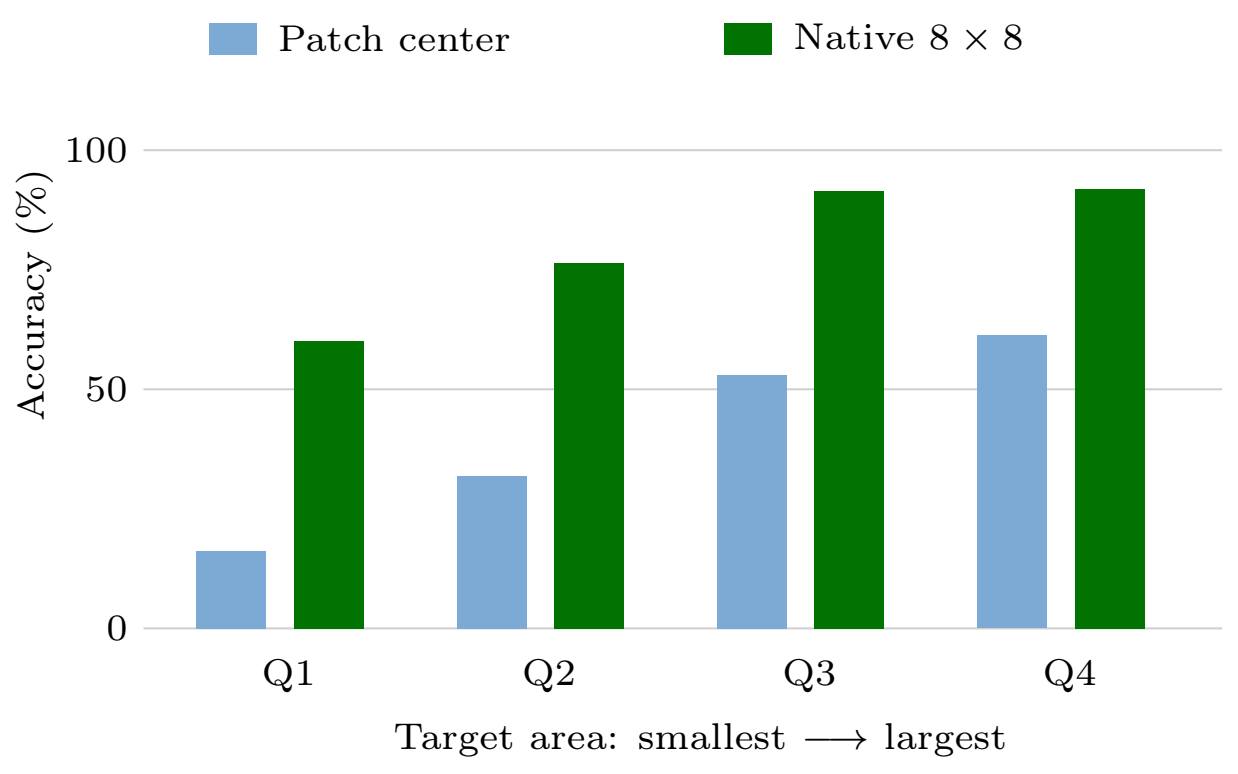
\begin{figure}[!ht]
\centering
\setlength{\unitlength}{1pt}
\begin{picture}(236,150)
\put(3,65){\rotatebox{90}{\scriptsize Accuracy (\%)}}
{\color{black!18}
\put(27,26){\line(1,0){203}}
\put(27,71){\line(1,0){203}}
\put(27,116){\line(1,0){203}}}
\put(12,23){\makebox(12,6)[r]{\scriptsize 0}}
\put(12,68){\makebox(12,6)[r]{\scriptsize 50}}
\put(8,113){\makebox(16,6)[r]{\scriptsize 100}}
{\color{cvprblue!65}
\put(37,26){\rule{13pt}{14.54pt}}
\put(86,26){\rule{13pt}{28.71pt}}
\put(135,26){\rule{13pt}{47.62pt}}
\put(184,26){\rule{13pt}{55.14pt}}}
{\color{green!45!black}
\put(52,26){\rule{13pt}{54.09pt}}
\put(101,26){\rule{13pt}{68.81pt}}
\put(150,26){\rule{13pt}{82.25pt}}
\put(199,26){\rule{13pt}{82.71pt}}}
\put(31,13){\makebox(40,8){\scriptsize Q1}}
\put(80,13){\makebox(40,8){\scriptsize Q2}}
\put(129,13){\makebox(40,8){\scriptsize Q3}}
\put(178,13){\makebox(40,8){\scriptsize Q4}}
\put(26,1){\makebox(205,8){\scriptsize Target area: smallest $\longrightarrow$ largest}}
\put(36,134){\textcolor{cvprblue!65}{\rule{9pt}{6pt}}}
\put(49,133){\makebox(69,8)[l]{\scriptsize Patch center}}
\put(133,134){\textcolor{green!45!black}{\rule{9pt}{6pt}}}
\put(146,133){\makebox(85,8)[l]{\scriptsize Native $8\times8$}}
\end{picture}
\caption{\textbf{Localization by target size on ScreenSpot-Pro.}
Within-patch refinement improves accuracy across all four target-area
quartiles, raising smallest-target accuracy from 16.16\% to 60.10\%.}
\label{fig:supp-target-size}
\end{figure}

\paragraph{Target-size analysis.}
We partition ScreenSpot-Pro targets into quartiles by normalized bounding-box
area. \Cref{fig:supp-target-size} shows that refinement improves accuracy
across all four size groups. In the smallest-target quartile, native
refinement raises accuracy from 16.16\% to 60.10\%, a 43.94-point gain.
By resolving positions within the selected patch, the offset readout
provides the subpatch precision needed for fine-grained GUI interaction.

%% file: supplement/representative_trajectories.tex
\clearpage
\section{OSWorld-JEV benchmark reference trajectories}
\label{sec:supp-representative-trajectories}

\noindent
These benchmark reference demonstrations illustrate type-free solutions
and preservation constraints across all six OSWorld-JEV categories.
Each trajectory uses type-free GUI actions; clipboard reuse, where present,
copies existing text. Each figure presents six sampled screenshot crops
in chronological order with their original step indices.
Captions summarize the task and its preservation constraints.

\medskip
\begingroup
\captionsetup{font=small,skip=6pt,hypcap=false}
\noindent\begin{minipage}{\textwidth}
\centering
\includegraphics[width=\linewidth]{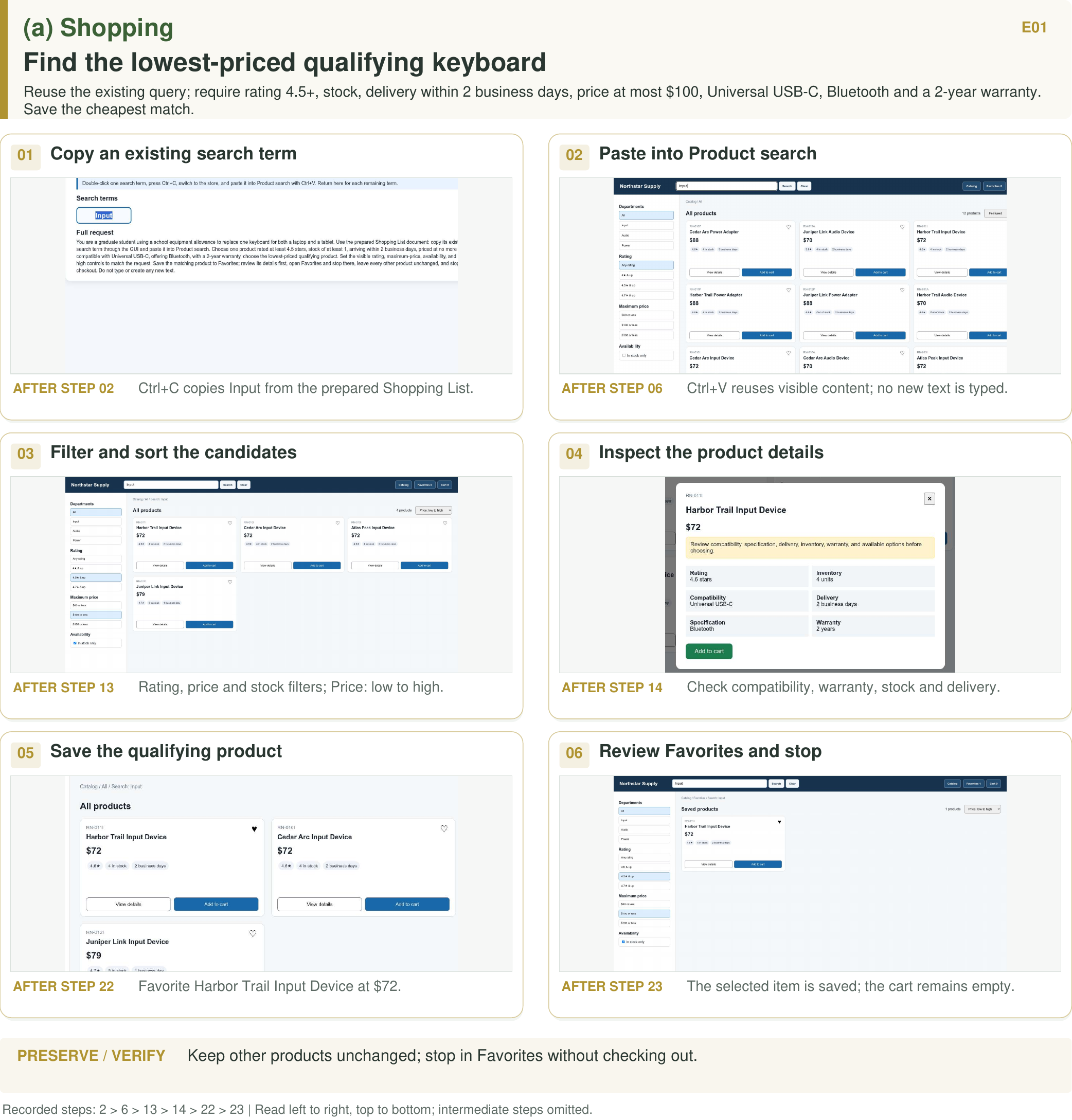}
\captionof{figure}{\textbf{Shopping: constraint-based product selection.}
The reference copies an existing search term, applies visible filters, checks
product specifications, and saves the qualifying item without entering new text.}
\label{fig:supp-traj-shopping}
\end{minipage}

\clearpage
\noindent\begin{minipage}{\textwidth}
\centering
\includegraphics[width=\linewidth]{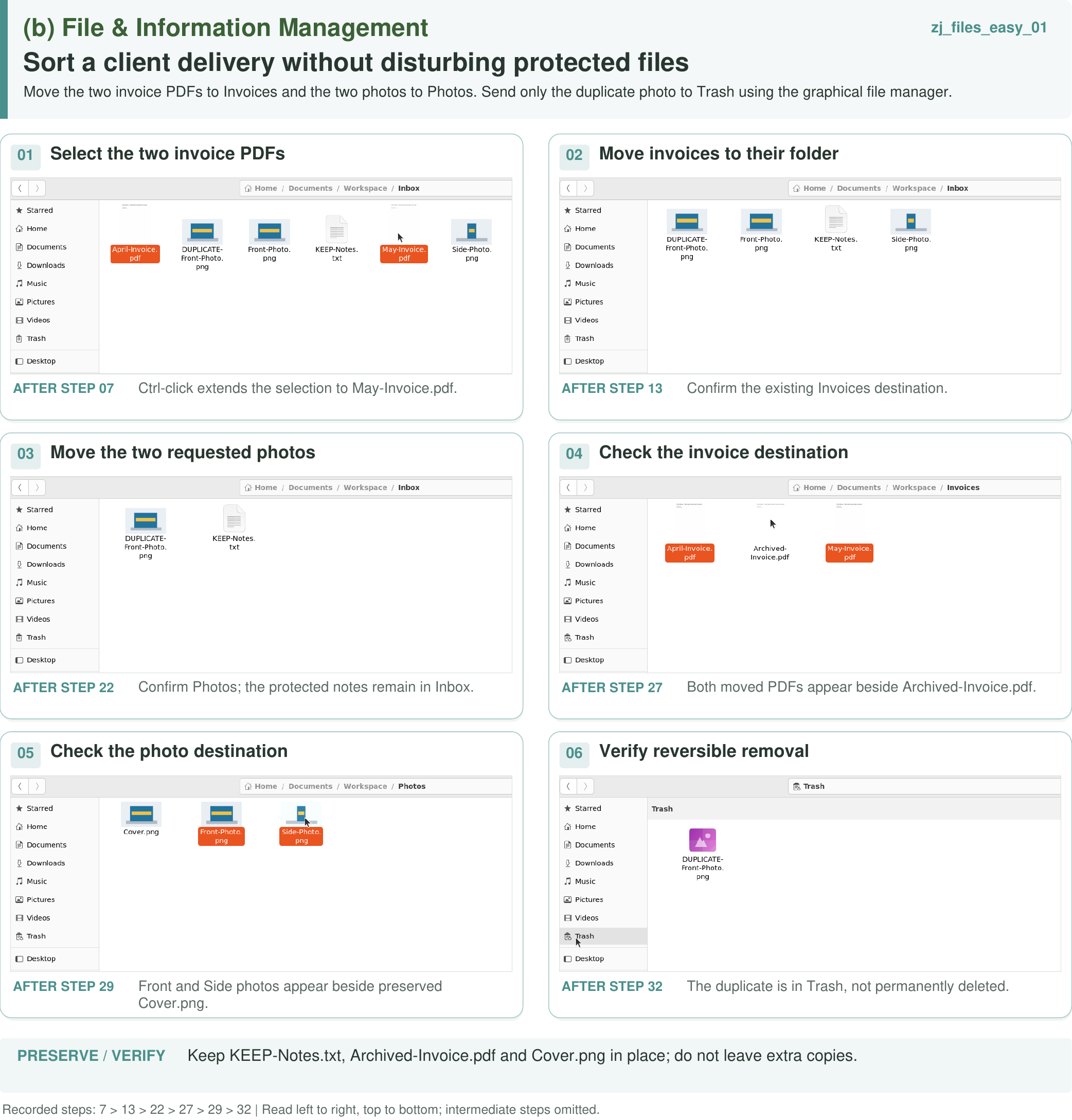}
\captionof{figure}{\textbf{File \& Information Management: selective file organization.}
GUI selection and move operations route invoices and photos to their existing
folders. Destination checks and Trash inspection expose both the requested
changes and preservation constraints.}
\label{fig:supp-traj-files}
\end{minipage}

\clearpage
\noindent\begin{minipage}{\textwidth}
\centering
\includegraphics[width=\linewidth]{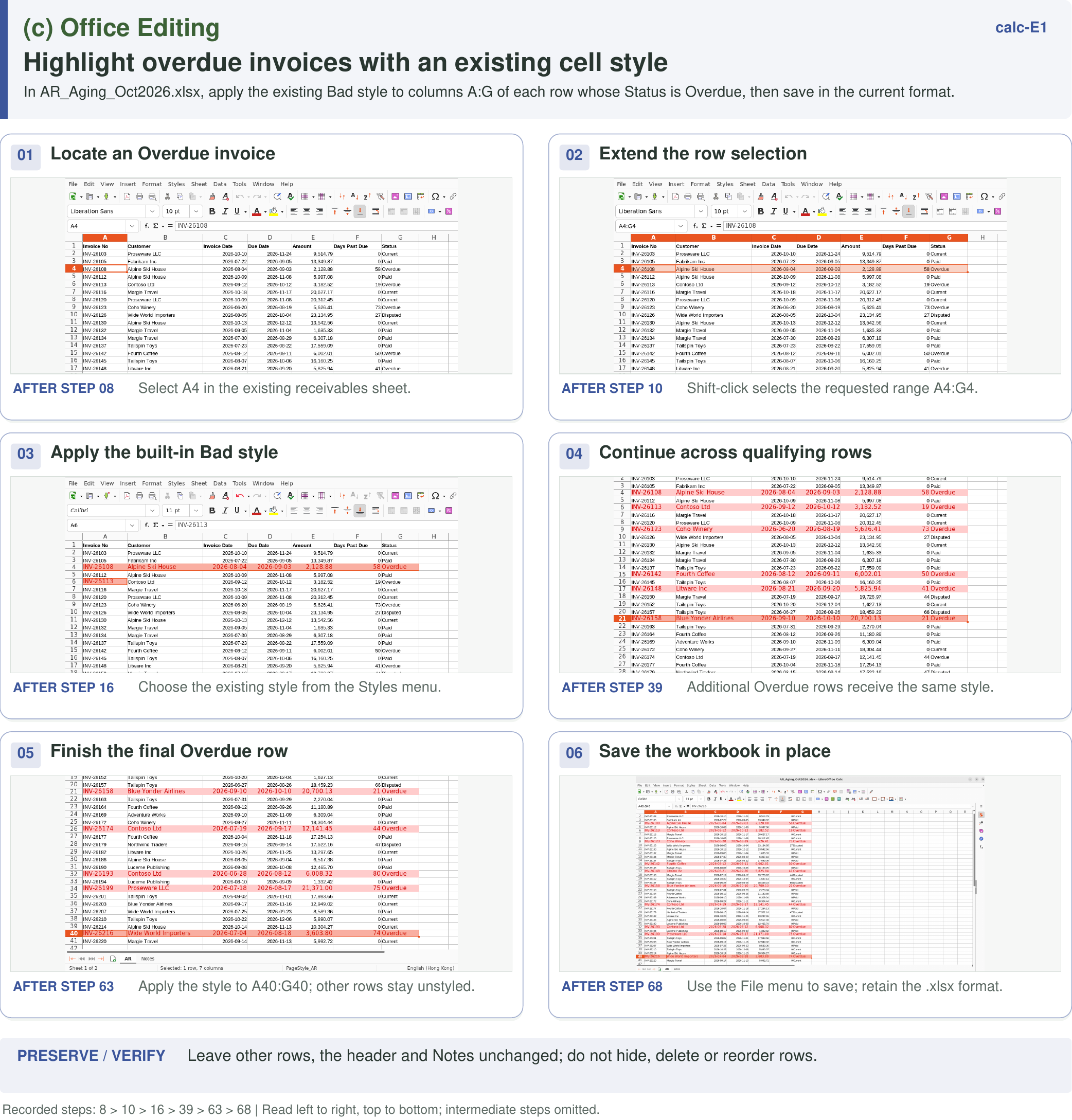}
\captionof{figure}{\textbf{Office Editing: selective spreadsheet formatting.}
The reference selects existing cell ranges, applies the built-in Bad style to
Overdue rows, and saves in place. Screenshots use the next recorded pre-action
observation, showing the state after each labeled action.}
\label{fig:supp-traj-office}
\end{minipage}

\clearpage
\noindent\begin{minipage}{\textwidth}
\centering
\includegraphics[width=\linewidth]{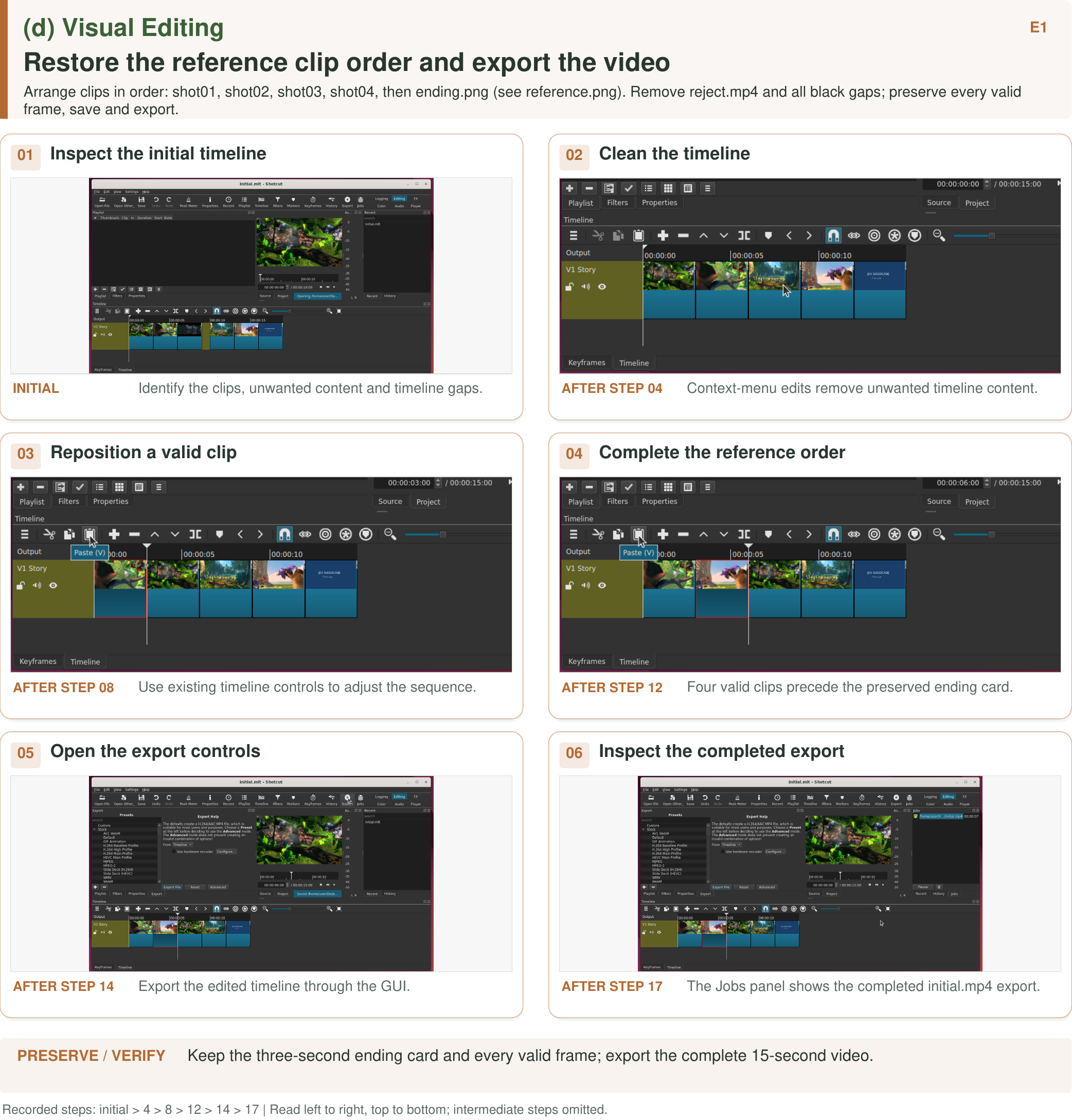}
\captionof{figure}{\textbf{Visual Editing: timeline editing in Shotcut.}
Sampled states show cleanup, clip repositioning, export controls, and the
completed export job. GUI operations preserve valid media while removing
unwanted timeline content.}
\label{fig:supp-traj-visual}
\end{minipage}

\clearpage
\noindent\begin{minipage}{\textwidth}
\centering
\includegraphics[width=\linewidth]{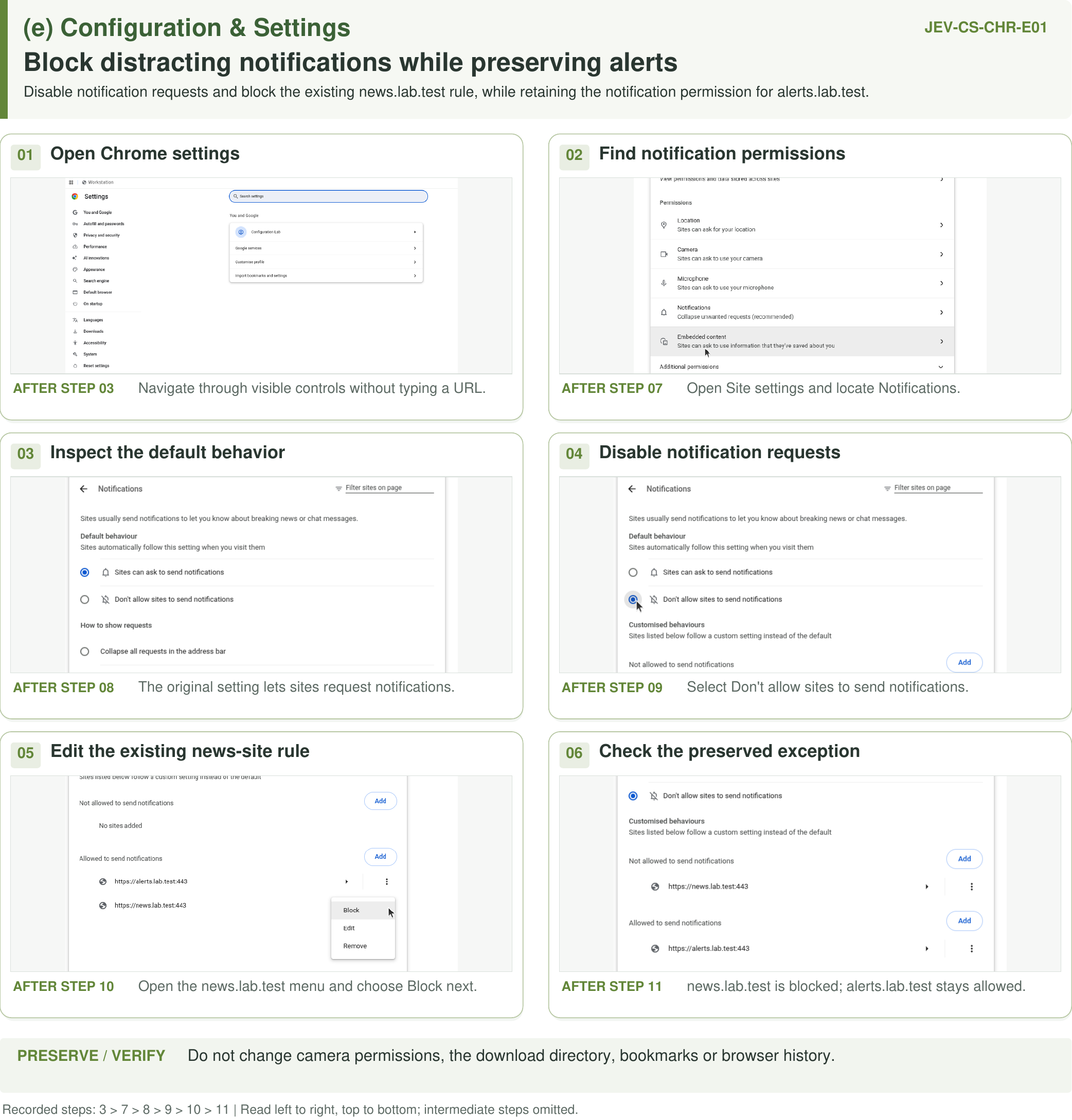}
\captionof{figure}{\textbf{Configuration \& Settings: targeted browser configuration.}
The reference changes the default notification policy and the existing news-site
rule while keeping the alert-site exception. Task wording is summarized in
English from the recorded instruction.}
\label{fig:supp-traj-settings}
\end{minipage}

\clearpage
\noindent\begin{minipage}{\textwidth}
\centering
\includegraphics[width=\linewidth]{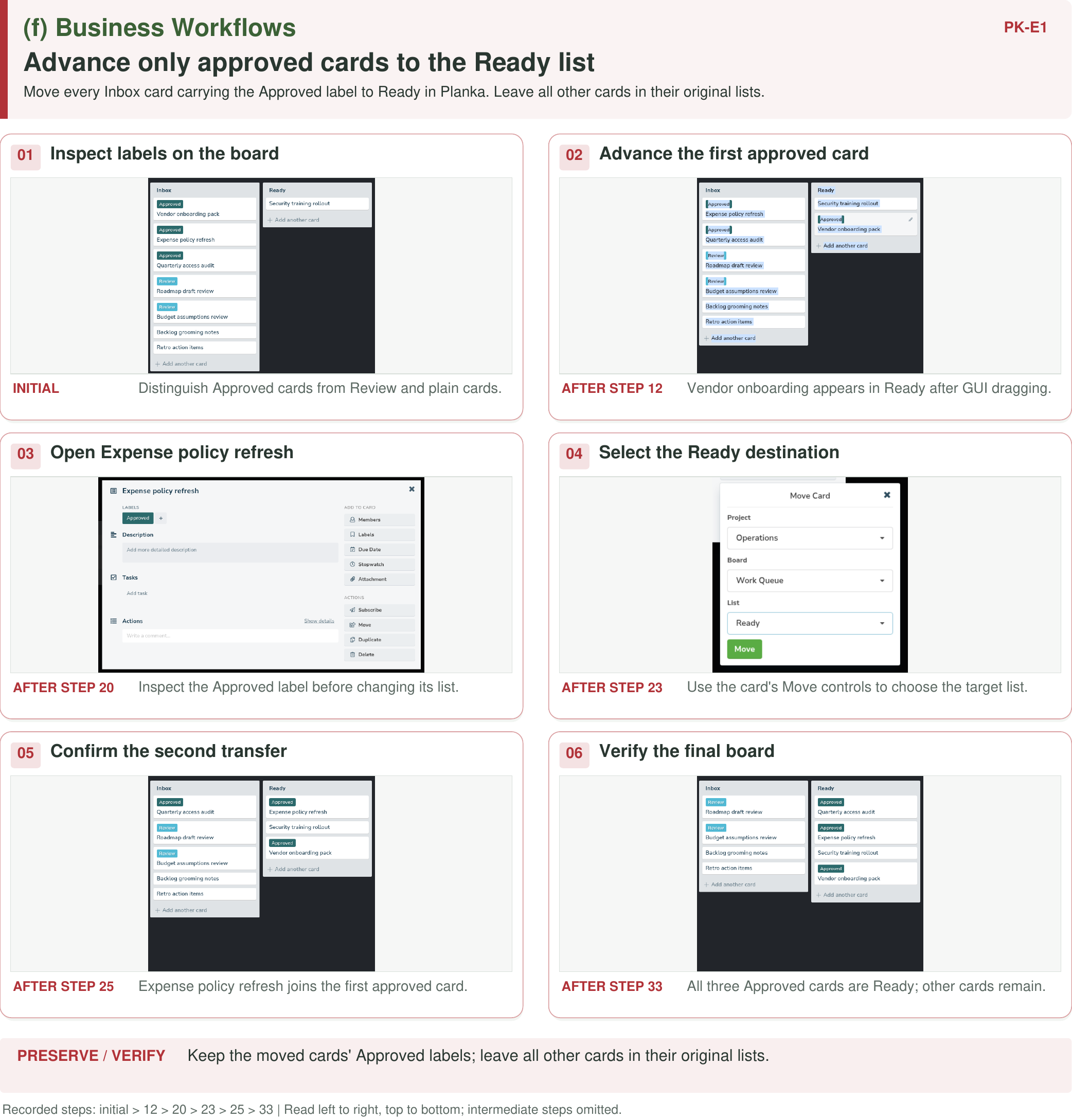}
\captionof{figure}{\textbf{Business Workflows: rule-governed state transitions in Planka.}
Only Approved cards advance from Inbox to Ready; all other cards remain in their
original lists. This example uses the accompanying business-flow golden reference
from the same dataset release.}
\label{fig:supp-traj-business}
\end{minipage}
\endgroup